\documentclass[10pt,journal,compsoc]{IEEEtran}

\ifCLASSOPTIONcompsoc
  \usepackage[nocompress]{cite}
\else
  \usepackage{cite}
\fi
\ifCLASSINFOpdf
\else
\fi
\usepackage{booktabs}
\usepackage{paralist} 
\usepackage{mathrsfs,amsfonts,amsmath,amsthm,amssymb}
\usepackage{multicol}
\usepackage{subfigure}
\usepackage{graphicx}
\usepackage{multirow}
\usepackage{subfigure}
\usepackage{algorithm}
\usepackage{algorithmic}
\usepackage{setspace}
\usepackage{siunitx}

\newtheorem{theorem}{Theorem}
\newtheorem{Lemma}{Lemma}

\theoremstyle{definition}
\newtheorem{definition}{Definition}

\theoremstyle{definition}

\theoremstyle{definition}
\newtheorem{remark}{Remark}

\theoremstyle{definition}

\usepackage{color}

\newcommand{\red}[1]{{\color{red}{#1}}}

\def \y {\mathbf{y}}

\def \w {\mathbf{w}}

\def \a {\mathbf{a}}

\def \c {\mathbf{c}}

\def \u {\mathbf{u}}
\def \v {\mathbf{v}}

\def \X {\mathbf{X}}

\def \B {\mathbf{B}}

\def \Z {\mathbf{Z}}
\def \G {\mathbf{G}}
\def \F {\mathbf{F}}

\def \V {\mathbf{V}}
\def \P {\mathbf{P}}
\def \S {\mathbf{S}}
\def \U {\mathbf{U}}
\def \I {\mathbf{I}}
\def \H {\mathbf{H}}
\def \A {\mathbf{A}}
\def \C {\mathbf{C}}
\def \D {\mathbf{D}}

\def \W {\mathbf{W}}
\def \Z {\mathbf{Z}}

\def \R {\mathbb{R}}

\def \E {\mathbb{E}}
\def \Pr {\mathbb{P}}
\def \Tr {\mathrm{Tr}}
\def \diag {\mathbb{D}}
\def \bS {\mathbf{\Sigma}}
\def \bT {\mathbf{\Phi}}
\def \bmu {\pmb{\mu}}
\def \bvarphi {\pmb{\varphi}}
\def \bvarsigma {\pmb{\varsigma}}

\def \l {\ell}

\usepackage{url}

\begin{document}
%
\title{Making Online Sketching Hashing Even Faster}
%
%
%
%

\author{Xixian Chen, 
Haiqin~Yang,~\IEEEmembership{Member,~IEEE,}
Shenglin~Zhao,~\IEEEmembership{Member,~IEEE,}
Irwin~King~\IEEEmembership{Fellow,~IEEE,}
and Michael~R.~Lyu,~\IEEEmembership{Fellow,~IEEE}
\thanks{
X.~Chen~(corresponding author) and S.~Zhao are with the Youtu Lab, Tencent,
Shenzhen, China, postal code: 518057, and also with the Department of Computer Science and Engineering, The Chinese University of Hong Kong, Hong Kong.  \{xixianchen, henryslzhao\}@tencent.com}
\thanks{
I.~King and M.~R.~Lyu, are with Shenzhen Research Institute, The Chinese University of Hong Kong, Shenzhen, China, postal code: 518057, and Department of Computer Science and  Engineering, The Chinese University of Hong Kong, Shatin, N.T., Hong Kong. \{king, lyu\}@cse.cuhk.edu.hk}
\thanks{H.~Yang is affiliated with Meitu (China), Hong Kong and Department of Computing, Hang Seng University of Hong Kong.  haiqin.yang@gmail.com.}
\thanks{The work was fully supported by the Research Grants Council of the Hong Kong Special Administrative Region, China (No. CUHK 14208815, No. CUHK 14234416, and Project No.~UGC/IDS14/16)}
}

%
%

\markboth{
IEEE Transactions on Knowledge and Data Engineering,~Vol.~xx, No.~x, xx~20xx}%
{IEEE Trans. Knowl. Data Eng.}
%



\IEEEtitleabstractindextext{%
\begin{abstract}
Data-dependent hashing methods have demonstrated good performance in various machine learning applications to learn a low-dimensional representation from the original data.  However, they still suffer from several obstacles: First, most of existing hashing methods are trained in a batch mode, yielding inefficiency for training streaming data.  Second, the computational cost and the memory consumption increase extraordinarily in the big data setting, which perplexes the training procedure.  Third, the lack of labeled data hinders the improvement of the model performance.  To address these difficulties, we utilize online sketching hashing (OSH) and present a FasteR Online Sketching Hashing (FROSH) algorithm to sketch the data in a more compact form via an independent transformation.  We provide theoretical justification to guarantee that our proposed FROSH consumes less time and achieves a comparable sketching precision under the same memory cost of OSH.  We also extend FROSH to its distributed implementation, namely DFROSH, to further reduce the training time cost of FROSH while deriving the theoretical bound of the sketching precision.  Finally, we conduct extensive experiments on both synthetic and real datasets to demonstrate the attractive merits of FROSH and DFROSH. 
\end{abstract}

}

\maketitle

\IEEEdisplaynontitleabstractindextext

%
\IEEEpeerreviewmaketitle

\begin{table*}[htp]
\centering \caption{Key notations used in this paper.\label{tab:notation}}
\begin{tabular}{p{.25\textwidth}p{.74\textwidth}}
\toprule
{\bf Notations} & {\bf Description} \\\midrule
${\bmu}$,  ${\bf W}$ & Bold small and capital letters denote vectors and matrices, respectively. \\\midrule
$[k]$ & An integer set consisting of $1, 2, \ldots, k$\\\midrule
$\a^i$ ($\a_j$, or $a_{ij}$) & The $i$-th row (the $j$-th column, or the $(i, j)$-th element) of $\A$, where $\a^i\in\R^{1\times d}$ for the matrix $\A\in \R^{n\times d}$ and $i\in [n]$, $j\in [d]$ \\\midrule
$\{\A_t\}_{t=1}^k$ & A set of $k$ matrices consisting of $\A_1, \A_2, \ldots, \A_k$ \\\midrule
$\a_{t,ij}$ ($\a_{t,i}$) & The $(i, j)$-th element (the $i$-th column) of matrix $\A_t$\\\midrule
$\A^T$ & The transpose of $\A$ \\\midrule
$\Tr (\A)$ & The trace of $\A$ \\\midrule
$|a|$ & The absolute value of a real number $a$ \\\midrule
$\|\A\|_2$ ($\|\A\|_F$) & The spectral (Frobenius) norm of $\A$\\\midrule
$\|\a\|_q=(\sum_{j=1}^d|a_j|^q)^{1/q}$ &  The $\l_q$-norm of $\a\in\R^d$, where $q\geq 1$
\\\midrule
$\diag(\A)$ & The square diagonal matrix whose main diagonal has only the main diagonal elements of $\A$ \\\midrule
\multirow{2}{*}{\begin{tabular}{@{}l@{}c@{}l@{}}
$\A$&=&$\U\bS\V^T=\sum_{i=1}^\rho \sigma_i\u_i\v_i^T$\\
&=&$\U_k\bS_k\V_k^T+\U_k^\perp\bS_k^\perp
{\V_k^{\perp}}^T$
\end{tabular}} & The SVD of $\A$, where $\mbox{rank}(\A)=\rho$, $\A_k:=\U_k\bS_k\V_k^T$ represents the best rank $k$ approximation to $\A$, and $\sigma_i(\A)$ denotes the $i$-th largest singular value of $\A$
\\\midrule
$\A\preceq \B$ & $\B-\A$ is positive semi-definite. 
\\\bottomrule
\end{tabular}
\end{table*}
\section{Introduction}\label{sec:FROH_intro}
\IEEEPARstart{H}{ashing} on features is an efficient tool to conduct approximate nearest neighbor searches for many machine learning applications such as large scale object retrieval~\cite{DBLP:conf/cvpr/JegouDSP10}, image classification~\cite{DBLP:conf/cvpr/SanchezP11}, fast object detection~\cite{DBLP:conf/cvpr/DeanRSSVY13}, and image matching~\cite{DBLP:journals/pami/ChumM10}, etc.  The goal of hashing algorithms is to learn a low-dimensional representation from the original data, or equivalently to construct a short sequence of bits, called hash code in a Hamming space~\cite{wang2016survey}, for fast computation in a common CPU.  Previously proposed hashing algorithms can be categorized into two streams, data-independent and data-dependent approaches.  Data-independent approaches, e.g., locality sensitive hashing (LSH) methods~\cite{andoni2015practical,charikar2002similarity,gionis1999similarity,indyk1998approximate,shrivastava2014asymmetric}, try to construct several hash functions based on random projection.  They can be quickly computed with theoretical guarantee.  However, to attain acceptable accuracy, the  data-independent hashing algorithms have to take a longer code length~\cite{shrivastava2014asymmetric}, which increases the computation cost.  Contrarily, data-dependent hashing approaches utilize the data distribution information and usually can achieve better performance with a shorter code length.  These approaches can be divided into unsupervised and supervised approaches.  Unsupervised approaches learn hash functions from data samples instead of randomly generated functions to maintain the distance in the Hamming space~{\cite{gong2011iterative,jiang2015scalable,leng2015online,liu2017ordinal,liu2016towards,liu2014discrete,mukherjee2015nmf,shen2015learning,weiss2009spectral,yu2014circulant}}.  Supervised approaches utilize the label information and can attain better performance than that of unsupervised ones~\cite{kang2016column,li2016feature,li2015two,lin2014fast,neyshabur2013power,shen2015supervised,wang2010semi}.  

However, data-dependent hashing approaches still suffer from several critical obstacles.  First, in real-world applications, data usually appear fluidly and can be processed only once~\cite{leng2015online,liberty2013simple}.  The batch trained hashing approaches become costly because the models need to be learned from scratch when new data appears~\cite{DBLP:conf/eccv/CakirHS18,DBLP:conf/cvpr/0003CBS18,DBLP:conf/nips/KulisD09}.  Second, the data can be in a huge volume and in an extremely large dimension~\cite{dhillon2013new}, which yields high computational cost and large space consumption and prohibits the training procedure~\cite{leng2015online,liberty2013simple}.  Though batch-trained unsupervised hashing techniques, such as Scalable Graph Hash (SGH)~\cite{jiang2015scalable} and Ordinal Constraint Hashing (OCH)~\cite{liu2017ordinal}, have overcome the space inefficiency by performing multiple passes over the data, they increase the overhead of disk IO operations and yield a major performance bottleneck~\cite{wu2016single}.  Third, the label information is usually scarce and noisy~\cite{song2015supervised,woolam2009lacking,xie2016unsupervised}.  Fourth, the data may be generated in a large amount of distributed servers.  It would be more efficient to train the models from the local data independently and to integrate them afterwards.  However, the distributed algorithm and its theoretical analysis are not provided in the literature yet.  

To tackle the above difficulties, we present a FasteR Online Sketching Hashing (FROSH) algorithm and its distributed implementation, namely DFROSH.  The proposed FROSH not only absorbs the advantages of online sketching hashing (OSH)~\cite{leng2015online}, i.e., data-dependent, space-efficient, and online unsupervised training, but also speeds up the sketching operations significantly.  Here, we consider the unsupervised hashing approach due to relieving the burden of labeling data and therefore overcome several popular online supervised hashing methods, e.g., online kernel hashing (OKH)~\cite{huang2013online}, online supervised hashing~\cite{cakir2015online}, adaptive online hashing (AOH)~\cite{cakir2015adaptive}, and online hashing with mutual information (MIHash)~\cite{DBLP:conf/iccv/CakirHBS17}.  


In sum, we highlight the contributions of our proposed FROSH algorithm in the following:
\begin{compactitem}
\item First, we present a FasteR Online Sketching Hashing (FROSH) to improve the efficiency of OSH.  The main trick is to develop a faster frequent direction (FFD) algorithm via utilizing the Subsampled Randomized Hadamard Transform (SRHT). 

\item Second, we devise a space economic implementation for the proposed FFD algorithm.  The crafty implementation reduces the space cost from $O(d^2)$ to $O(d\l)$, attaining the same space cost of FD, where $d$ is the feature size and $\l$ is the sketching size with $\l<d$.

\item Third, we derive rigorous theoretical analysis of the error bound of FROSH and show that under the same sketching precision, our proposed FROSH consumes significantly less computational time, $\widetilde{O}(n\l^2+nd+d\l^2)$, than that of OSH with $O(nd\l + d\l^2)$, where $n$ is the number of samples.
 
\item Fourth, we propose a distributed implementation of FROSH, namely DFROSH, to further speed up the training of FROSH.  Both theoretical justification and empirical evaluation are provided to demonstrate the superiority of DFROSH.
\end{compactitem}  

The remainder of the paper is structured as follows.  In Section 2, we define the problem and review existing online sketching hashing methods.  In Section 3, we present our proposed FROSH, its distributed implementation, and detailed theoretical analysis.  In Section 4, we conduct extensive empirical evaluation and detail the results.  Finally, in Section 5, we conclude the whole paper. 

\section{Problem Definition and Related Work}

\subsection{Notations and Problem Definition} 
To make the notations consistent throughout the whole paper, we present some important notations with the specific meaning defined in Table~\ref{tab:notation}.  Given $n$ data in $d$ dimension, i.e., $\A\in\R^{n\times d}$, the goal of hashing is to seek a projection matrix $\W\in\R^{d\times r}$ for constructing $r$ hash functions to project each data point $\a^i\in\R^{1\times d}$ defined as follows: 
\begin{align}
h_k(\a^i)=\text{sgn}((\a^i-\bmu)\w_k),~~k=1,\ldots, r, 
\end{align}
where $\bmu$ is the row center of ${\A}$ defined by $\frac{1}{n}\sum_{i=1}^n\a^i$.  To produce an efficient code in which the variance of each bit is maximized and the bits are pairwise uncorrelated, one can maximize the following function~\cite{gong2011iterative}:
\begin{align}\label{eq:objective}
&\max_{\W}~~ \sum_{k=1}^rVar(h_k(\a)), \notag \\
\text{s.t.}~\E[h_{k_i}(\a)h_{k_j}&(\a)]=1~\text{for}~ i=j, ~\text{and}~ 0 ~\text{otherwise}. 
\end{align}
Adopting the same signed magnitude relaxation in~\cite{wang2010semi}, the objective function becomes~\cite{leng2015online}: 
\begin{align}\label{eq:objective}
\max_{\W} &~~  \Tr(\W^T(\A-\bmu)^T(\A-\bmu)\W),~   \text{s.t.}~\W^T\W=\I_r, 
\end{align}
where $(\A-\bmu)$ denotes a matrix of  $[\a^1-\bmu;\a^2-\bmu;\ldots;\a^n-\bmu]$. 

In Eq.~(\ref{eq:objective}), we only need to performs Principal Component Analysis (PCA) on $\A$, and $\W\in\R^{d\times r}$ results from the top $r$ right singular vectors of the covariance matrix $(\A-\bmu)^T(\A-\bmu)$.  When $d<n$, it requires $O(nd^2)$ time cost and $O(nd)$ space, which is infeasible for large $n$ and $d$~\cite{jolliffe2002principal}.

\subsection{Online Sketching Hashing}
\label{subsection:osh}
To reduce the computational cost of seeking $\W$, researchers have proposed various methods, e.g., random projection, hashing, and sketching, and provided the corresponding theoretical guarantee on the approximation~\cite{leng2015online,DBLP:journals/ftml/Mahoney11,wang2016survey}.  Among them, sketching is one of the most efficient and economic ways by only selecting a significantly smaller data to maintain the main information of the original data while guaranteeing the approximation precision~\cite{avron2014subspace,choromanska2016binary,liberty2013simple,luo2016efficient,woodruff2014sketching,chen2015training,chen2017toward}.


Online Sketch Hashing (OSH)~\cite{leng2015online} is an efficient hashing algorithm to solve the PCA problem of Eq.~(\ref{eq:objective}) in the online mode.  Given $n$ data points appearing sequentially, denoted by a matrix $\A\in\R^{n\times d}$, the goal of OSH is to efficiently attain a small mapping  matrix $\W^T\in\R^{r\times d}$ via constructing SVD on a small matrix $\B\in\R^{\l\times d}$, where $r$ is the number of hashing bits and $\l$ is the sketching size with $\l<d$; see step~13 in Algo.~\ref{alg:frosh}.  The key of OSH is to construct $\B$ from the centered data $(\A-\bmu)$ by the \textit{frequent direction} (FD) algorithm along with the creative idea of \textit{online centering procedure}~\cite{DBLP:conf/uai/ChenKL17,leng2015online}, i.e., the same procedure of steps~3 to 10 in Algo.~\ref{alg:frosh} except that the sketching method \textit{faster frequent direction} (FFD) in steps~4 and 8 is replaced by FD.

\begin{algorithm}[htbp]  
\caption{FasteR Online Sketching Hashing (FROSH)}
\label{alg:frosh}   
\begin{algorithmic}[1] 
\REQUIRE 
Data $\A=\{\A_j\in\R^{h_j\times d}\}_{j=1}^s$, sketching size $\l< d$, positive integer $\eta$, hashing bits $r$
\STATE Initialize sketching matrix with $\B = \mathbf{0}^{\l\times d}$ \label{frosh1}
\STATE Set $\bmu_1$ to be the row mean vector of $\A_1$
\STATE Let $\bvarphi=\bmu_1$,  $\tau=h_1$, $\xi=0$\label{step:frosh:cs}
\STATE Invoke FFD($\G_1$, $\B$), where $\G_1=(\A_1-\bmu_1)\in\R^{h_1\times d}$, $\B\in\R^{\l\times d}$ \quad\# Sketch $\G_1$ into $\B$ via FFD\label{step:frosh:ffd_1}
\FOR {$j\in\{2, \ldots, s\}$}
\STATE Set $\bmu_j$ to be the row mean vector of $\A_j$
\STATE Set $\bvarsigma=\sqrt{\frac{\tau h_j}{\tau+h_j}}(\bmu_j-\bvarphi)\in\R^{1\times d}$
\STATE Invoke FFD($\G_j$, $\B$), where $\G_j=[(\A_j-\bmu_j); \bvarsigma] \in\R^{(h_j+1)\times d}$ \quad\# Sketch $\G_j$ into $\B$ via FFD\label{step:frosh:ffd_2}
\STATE Set $\bvarphi=\frac{\tau\bvarphi}{\tau+h_j}+\frac{h_j\bmu_j}{\tau+h_j}$ \quad\# Update the row mean vector
\STATE Set $\tau=\tau+h_j$ \quad\# Update the number of data instances\label{step:frosh:ce} 
\STATE Set $\xi=\xi+1$
\IF {$\xi == \eta$}
\STATE Run the SVD of $\B\in\R^{\l\times d}$, and set the top $r$ right singular vectors as $\W^T\in\R^{r\times d}$ \label{step:frosh:svd}
\STATE Set $\xi=0$ 
\RETURN $\W$
\ENDIF
\ENDFOR
\end{algorithmic}  
\end{algorithm}

\section{Our Proposal}

\subsection{Motivation and FROSH}
It is noted that OSH consumes $O(nd\l+d\l^2)$ time with $O(nd\l)$ for sketching $n$ incoming data points and $O(d\l^2)$ for compute PCA on $\B$, while maintaining an economic storage cost at $O(d\l)$. This is still computationally expensive when $1\ll d \ll n$~\cite{dhillon2013new}.

By further reducing the sketching cost in OSH, we propose our FROSH algorithm, outlined in Algo.~\ref{alg:frosh} via utilizing a novel designed {\em Faster Frequent Directions} (FFD) algorithm.
\begin{remark}\label{remark1}
We elaborate more details about Algo.~\ref{alg:frosh}: 
\begin{compactitem}
\item Steps~\ref{step:frosh:cs}-\ref{step:frosh:ce} are the creative \textit{online centering procedure} proposed in OSH~\cite{leng2015online}, which guarantees $\G_{[j]}^T\G_{[j]}=(\A_{[j]}-\hat{\bmu}_j)^T(\A_{[j]}-\hat{\bmu}_j)$,  after the $j$-th iteration, where $\A_{[j]}=[\A_1;\A_2;\cdots;\A_j]$ are stacked vertically for all sequential sample $\A_j$ and  $\hat{\bmu}_j$ denotes the row mean vector of $\A_{[j]}$.  Via invoking $\rm{FFD}(\cdot,\cdot)$, we expect in the final step, $\B^T\B\approx \G_{[s]}^T\G_{[s]}$, which is equal to  $(\A_{[s]}-\hat{\bmu}_s)^T(\A_{[s]}-\hat{\bmu}_s)$, i.e., $(\A-\bmu)^T(\A-\bmu)$.  
\item  Step~\ref{step:frosh:ffd_1} and step~\ref{step:frosh:ffd_2} invoke FFD to sketch a smaller matrix $\B$ such that $\B^T\B \approx \G_{[j]}^T\G_{[j]}$ at the $j$-th iteration.  The key contribution of our FROSH is to design and utilize a faster sketching method in Algo.~\ref{alg:ffd} rather than the original FD.
\item\label{rem:rotation} The step~\ref{step:frosh:svd} is to compute the top $r$ right singular vectors $\W$, which yields $O(d\l^2)$ computational cost.
\item Here, we highlight again the advantages of FFD for the streaming data, which borrows the idea of FD in~\cite{liberty2013simple} and OSH~\cite{leng2015online}.  For example, regarding two (or more) data chunks, $\G_1\in\R^{h_1\times d}$ and $\G_2\in\R^{(h_2+1)\times d}$, we run FFD on $\G_1$ with $\B$ and achieve a sketching matrix $\B=\B_1\in\R^{\l\times d}$.  Relying on the current $\B=\B_1$, we then invoke FFD on $\G_2$ to make an update for $\B$ and obtain $\B=\B_2\in\R^{\l\times d}$.  This procedure is equivalent to directly invoking FFD once on $[\G_1;\G_2]\in\R^{(h_1+h_2+1)\times d}$ and results the same $\B_2$ for the final $\B$.
\item In sum, the total time cost of FROSH is $O(\mathrm{T}(\mathrm{FFD})+d\l^2)$, where  $\mathrm{T}(\mathrm{FFD}) = O(n\l^2+nd)$ for FFD as depicted in Remark~\ref{remk:fft_time_space}.
\end{compactitem}
\end{remark}

\begin{algorithm}[htbp]  
\caption{Faster Frequent Directions (FFD): FFD\,($\A$, $\B$), $\mbox{where}~\A~\in~\R^{n\times d},~\B~\in~\R^{\l\times d}$}
\label{alg:ffd}   
\begin{algorithmic}[1] 
\IF {$\B$ not exists}
\STATE Set $t=1$, $\B = \mathbf{0}^{\l\times d}$
\STATE Set $\F=\mathbf{0}^{m\times d}$ with $m=\Theta(d)$ \# Only need for notation
\ENDIF
\FOR {$i\in [n]$}
\STATE Insert $\a^i$ into {a zero valued row} of $\F$ \label{step:ffd:da1}
\IF {$\F$ has no zero value row}
\STATE Construct $\bT_t=\S_t\H\D_t\in\R^{(\l/2)\times m}$ at the $t$-th trial
\STATE Insert $\bT_t\F\in\R^{(\l/2)\times d}$ into the last $\frac{\l}{2}$ rows of $\B$ \label{step:ffd:da2}
\STATE $[\U,\bS,\V]=\textsc{SVD}(\B)$ \label{step:ffd:svd1}
\STATE $\widehat{\bS}=\sqrt{\max(\bS^2-\sigma_{\l/2}^2\I_l,0)}$ \label{step:ffd:svd2}
\STATE Set $\B=\widehat{\bS}\V^T$ \label{step:ffd:B}
\STATE Let $\B_t=\B$, $\C_t=\bT_t\F$ \# Only need for proof notations
\STATE Set $\F=\mathbf{0}^{m\times d}$, $t=t+1$
\ENDIF
\ENDFOR
\end{algorithmic}  
\end{algorithm}

\subsection{Fast Frequent Directions}
Our proposed FFD algorithm is outlined in Algo.~\ref{alg:ffd}.  
\begin{remark}
Here, we emphasize on several key remarks:
\begin{compactitem}
\item Step~\ref{step:ffd:da1}-step~\ref{step:ffd:da2} are the core steps of FFD: it first collects $m$ data points sequentially and stores them in $\F\in\R^{m\times d}$ with squeezing all zero valued rows; next, it constructs a new Subsampled Randomized Hadamard Transform (SRHT) matrix, $\bT=\S\H\D\in\R^{\l/2\times m}$ (the subscript $t$ indicates that $\S_t$ and $\D_t$ in step 8 are drawn independently at different trials), and 

$\bullet~\S\in \R^{q\times m}$: a scaled randomized matrix with its each row uniformly sampled \textit{without replacement} from $m$ rows of the ${m\times m}$ identity matrix rescaled by $\sqrt{\frac{m}{q}}$; 
$\bullet~\D$: an $m\times m$ diagonal matrix with the elements as i.i.d. Rademacher random variables (i.e., $1$ or $-1$ in an equal probability); 

$\bullet~\H\in\{\pm 1\}^{m\times m}$: a Hadamard matrix defined by $h_{ij} = (-1)^{\langle i-1, j-1\rangle}$, where $\langle i-1, j-1\rangle$ is the dot-product of the $b$-bit binary vectors of the integers $i-1$ and $j-1$, $b= \min\{\left\lceil \log (i+1) \right\rceil, \left\lceil \log (j+1)\right\rceil\}$, and $\left\lceil x\right\rceil$ returns  the least integer that is greater than or equal to $x$.

The Hadamard matrix can also be recursively defined by 
\begin{equation}
\H_m=\begin{bmatrix}
\H_{m/2} & \H_{m/2}\\
\H_{m/2} & -\H_{m/2}
\end{bmatrix}\quad
\text{and} \qquad
\H_2=\begin{bmatrix}
1 & 1 \\
1 & -1 \notag
\end{bmatrix}, 
\end{equation}
where $m$ is the size of the matrix.  The normalized Hadamard matrix is denoted by $\H=\sqrt{\frac{1}{m}}\H_m$.  Due to the recursive structure of $\H_m$, for $\H_m\a$, we only take $O(m\log m)$ time to compute it and $O(m)$ space to store it~\cite{ailon2009fast}, respectively.

The objective of computing $\bT\F$ is to compress $\F$ from the size of $m$ to $\l/2$ and finally yield a new matrix ${\l\times d}$ matrix $\B$ by concatenating the newly compressed data and the previously shrunk data via conducting step~\ref{step:ffd:da2}.

\item After constructing $\B$, FFD conducts SVD on $\B$ and condenses the original information in the first ${\l\over 2}$ rows of $\B$ in the steps of~\ref{step:ffd:svd1} and~\ref{step:ffd:svd2}.  This procedure is the same as that of FD in OSH and allows new data to be concatenated into the last ${\l\over 2}$ rows of $\B$ in the next iteration.  
\item The step of~\ref{step:ffd:B} is to reconstruct $\B$, which condenses the original information in the first ${\l\over 2}$ rows of $\B$.  This makes it effective and allows that in the step of~\ref{step:ffd:da2}, $\bT_t\F$ can only be concatenated into the last ${\l\over 2}$ rows of $\B$.
\item\label{remk:fft_time_space} In sum, at each iteration, the time cost for step~\ref{step:ffd:da2} is $\widetilde{O}(md)$ and it costs $O(d\l^2)$ for conducting SVD from step~\ref{step:ffd:svd1}-\ref{step:ffd:svd2}.  Totally, there are $O(n/m)$ iterations and yield the time cost of $\widetilde{O}({n\over m}d\l^2+{n\over m}md)=\widetilde{O}(n\l^2\frac{d}{m}+{n}d)$ for running Algo.~\ref{alg:ffd}.  The time cost becomes $\widetilde{O}(n\l^2+nd)$ when $m=\Theta(d)$, which is smaller than $O(nd\l)$ in FD of OSH when $\l \ll d$.
\end{compactitem}
\end{remark}

\subsection{Space-efficient Implementation of FFD}\label{subsection:imple}

Though the SRHT operation can improve the efficiency of sketching, it needs $O(md)$ memory to store the $m\times d$ matrix $\F$.  This yields $O(d^2)$ space consumption when $m=\Theta(d)$.  It is practically prohibitive because the storage space can be severely limited in real-world applications~\cite{liberty2013simple}.  {To make FFD competitive to the original FD algorithm in space cost, we design a crafty way to compute the data progressively and yield the same space cost as FD.}  

\begin{figure}[htbp]
\centering
\subfigure{
\begin{minipage}[b]{0.48\textwidth} 
\includegraphics[width=1.0\textwidth]{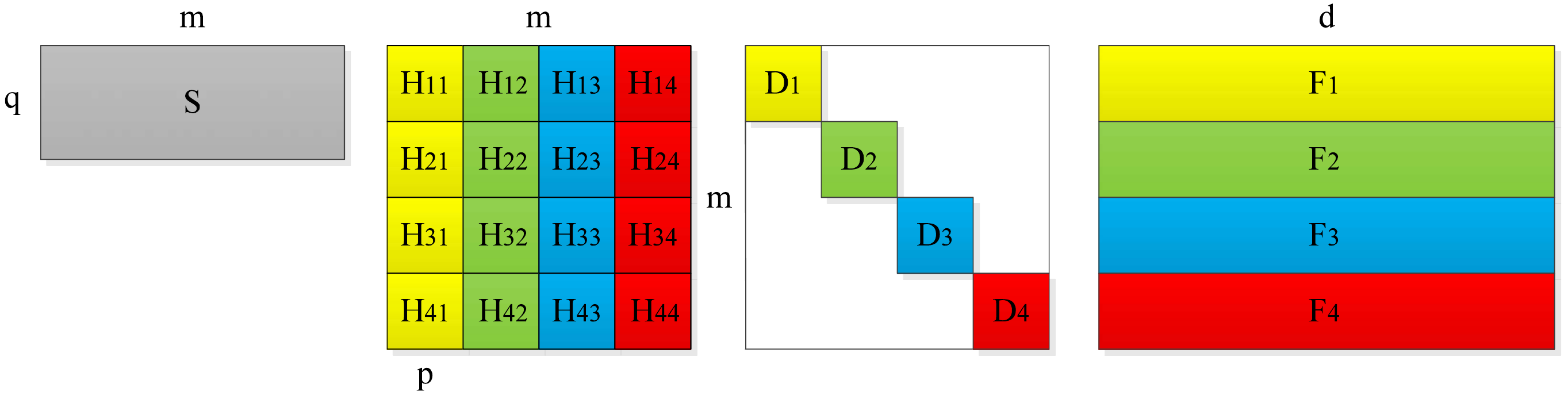} 
\end{minipage}
}
\caption{Space-efficient implementation of $\bT\F=\S\H\D\F$.  To simplify the illustration, we set ${m}$ to $2^c~p$, where $c=2$.  }
\label{fig:imple}
\end{figure}

Figure~\ref{fig:imple} illustrates our proposed space-efficient implementation of FFD.  For simplicity, we assume $m=2^b$, where $b$ is a positive integer.  We can always find $p$ in $[{q\over 2}, q]$ such that $m =2^c~p$ and $c$ is a positive integer.  We then divide the ${m\times d}$ data matrix $\F$ into $2^c$ blocks denoted by $\{\F_i\in\R^{p\times d}\}_{i=1}^{2^c}$.  Regarding $\bT$, the diagonal $m\times m$ matrix $\D$ can be divided into $2^c$ square blocks denoted by $\{\D_i\in\R^{p\times d}\}_{i=1}^{2^c}$.  The Hadamard matrix $\H$ can be also divided into $(2^c)^2=4^c$ square blocks denoted by $\{\H_{ij}\in\R^{p\times p}\}_{i,j=1}^{2^c}$.  

\begin{remark}
The space-efficient implementation of FFD can then be easily achieved:
\begin{compactitem}
\item\label{eq:F1_1} For a mini-batch incoming data $\F_1\in\R^{p\times d}$, the space cost is $O(pd)$.  By computing $\bT\F_1$, we will attain a ${q\times d}$ matrix.  Hence, the space cost is $O(pd+qd)=O(dq)$.  

\item\label{eq:F1_2} To compute $\bT\F_1$, i.e., $\S[\H_{11};\H_{21};\cdots;\H_{2^c1}]\D_1\F_1$, we first compute $\Z=\{\Z_i\}_{i=1}^{2^c}$, where $\Z_i=\{\H_{i1}\D_1\F_1\}$.  Picking any element, say $i=1$, the computational cost of $\Z_1=\H_{11}\D_1\F_1\in\R^{p\times d}$ is $O(pd\log p)$.  For other $\Z_i$, $i\neq 1$, due to the simple structure of the Hadamard matrx, i.e., $h_{ij} = (-1)^{\langle i-1, j-1\rangle}$, we have $\Z_i=\Z_1$ or $\Z_i=-\Z_1$.  A smart trick of computing $\Z_i$ is to check the first entry of each $\H_{i1}$ in the Hadamard matrix because $\H_{i1}=+\H_{11}$ or $\H_{i1}=-\H_{11}$, which yields $O(\log m)$ additional time cost. 

\item The second step of computing $\bT\F_1$ is to compute $\S\Z$, which is equivalent to selecting at most $q$ rows from $\Z$.  The time cost is negligible.  Hence, the total time cost of computing $\bT\F_1$ is $O(pd\log p)+O(q\log m)+O(qd)$.  The first term is to compute $\Z_1$.  The second term is to determine the sign difference of $\Z_1$ and $\Z_i$,  where $i=2, \ldots, 2^c$.  Since only $q$ rows are sampled from $\Z_i$, the total number of the focused $\Z_i$ is $\min(2^c-1, q)\leq q$.  The third term is to compute at most $q$ rows in $\{\Z_i\}_{i=2}^{2^c}$ that are sampled by $\S$.  

\item This procedure can be continuously enumerated from $\F_2$ to $\F_{2^c}$, where the time cost and the space cost consume asymptotically unchanged as computing $\bT\F_1$.  Hence, the total time cost of computing $\bT\F$ is $2^c[O(pd\log p)+O(q\log m)+O(qd)]=O(md\log q)$ because $m$ is usually set to $O(d)$ for the practical usage and therefore $\log m\leq O(d\log q)$.  The space cost of $\bT\F$ is $O(pd+qd)=O(qd)$.  Practically, we set $q=\l/2$ and obtain the space cost of $O(d\l)$ and the time cost of $O(md\log \l)$.  Hence, we maintain the time cost of the original FFD, i.e., $O(n\l^2+nd)$, significantly reducing from $O(nd\l+d\l^2)$ in FD, while reducing its space cost from $O(d^2)$ to $O(d\l)$, which is the same space cost of FD.
\end{compactitem}
\end{remark}

\subsection{Distributed FROSH}\label{sec:distributed_frosh}
In real-world applications, data may be stored locally in different servers.  To accelerate the training of FROSH, we propose a distribution implementation of FROSH, namely DFROSH, sketched in Algorithm~\ref{alg:DFROSH}, where we assume data  $\A=\{\A_i\}_{i=1}^\omega$ are stored in $\omega$ machines.

\begin{remark}
Here, we highlight several key remarks:
\begin{compactitem}
\item Step~\ref{step:disosh:dis} asynchronously invokes FROSH in $\omega$ machines independently.  
\item Step~\ref{step:DFROSH_Step1} to step~\ref{step:disosh:mean2} are to concatenate the sketched results from $\omega$ machines and invoke FD to sketch the concatenated matrix.  The time cost is negligible because the size of the concatenated sketched data is very small.  In step~\ref{step:center}, the online centering procedure is conducted and the objective is the same as that in Remark~\ref{remark1} to ensure that we can sketch on a certain matrix, saying $\widehat{\G}$, such that $\widehat{\G}^T\widehat{\G}=(\A-\bmu)^T(\A-\bmu)$, where $\A=[\A_1;\A_2;\cdots;\A_\omega]$. 

\item {It is noted that the sketching precision of DFROSH is the same as the one sketching the concatenated data throughout all the distributed machines, which means that the sketching precision of DFROSH and FROSH is equivalent; see detailed theoretical analysis in Theorem~\ref{thm:DFROSH}.  In terms of the time cost, step~\ref{step:DFROSH_Step1} to step~\ref{step:disosh:mean2} is negligible since $\B_i$ and $\omega$ are small, and step~\ref{step:disosh:dis} takes only about $1/\omega$ time cost of that running FROSH on $\A=[\A_1;\A_2;\cdots;\A_\omega]$ in a single machine with conducting online data centering procedure.}
\end{compactitem}
\end{remark}

\begin{algorithm}[htbp]  
\caption{Distributed FROSH (DFROSH): DFROSH\,($\{\A_i\}_{i=1}^\omega$, $\B$), $\mbox{where}~\A_i~\in~\R^{n_i\times d},~\B~\in~\R^{\l\times d}$}
\label{alg:DFROSH}   
\begin{algorithmic}[1] 
\FOR {$i\in [\omega]$}
\STATE Obtain the sketch $\B_i\in\R^{\l\times d}$, row mean vector $\bmu_i\in\R^{1\times d}$, and data size $n_i$ of $\A_i$ by running steps~\ref{frosh1} to~\ref{step:frosh:ce} of FROSH for $\A_i$ \quad\# In the $i$-th distributed machine \label{step:disosh:dis}
\ENDFOR
\STATE \label{step:DFROSH_Step1} Receive $\B_1$, $\bmu_1$, and $n_1$
\STATE \label{step:bigB1} Let $\bvarphi=\bmu_1$,  $\tau=n_1$ \label{step:disosh:mean1}
\STATE Initialize sketched matrix by $\B = \B_1^{\l\times d}$ \# In the center machine
\FOR {$i\in\{2, \ldots, \omega\}$}
\STATE Receive $\B_i$, $\bmu_i$, and $n_i$
\STATE\label{step:center} Set $\bvarsigma=\sqrt{\frac{\tau n_i}{\tau+n_i}}(\bmu_i-\bvarphi)\in\R^{1\times d}$
\STATE Invoke FD($\G_i$, $\B$), where $\G_i=[\B_i; \bvarsigma] \in\R^{(n_i+1)\times d}$ \quad\# Sketch $\G_i$ into $\B$ by FD\label{step:disosh:s21}
\STATE Set $\bvarphi=\frac{\tau\bvarphi}{\tau+n_i}+\frac{n_i\bmu_i}{\tau+n_i}$ \quad
\STATE Set $\tau=\tau+n_i$ \quad \label{step:disosh:ce}
\ENDFOR \label{step:disosh:mean2}
\STATE Compute the SVD of $\B\in\R^{\l\times d}$ and assign the top $r$ right singular vectors as $\W^T\in\R^{r\times d}$ \label{step:disosh:svd1}
\RETURN $\W$
\end{algorithmic}  
\end{algorithm}


\subsection{Analysis}\label{sec:frosh_analysis}

Before proceeding the theoretical analysis, we present the following definitions:
\begin{definition}\label{def:theorem_notations}
Without loss of generality, we let $q$ be $\l/2$ and $p$ be ${n\over m}$, the number of times to proceed step 9 in Algo.~\ref{alg:ffd}.  Let the input be  $\A=[\A_1;\A_2;\cdots;\A_p]\in\R^{n\times d}$ with $\{\A_t\in\R^{m\times d}\}_{t=1}^p$, where a mild but practical assumption has also been made as $\lambda_1\leq \|\A_t\|_F^2\leq \lambda_2$ with $\lambda_1$ and $\lambda_2$ close to each other~\cite{DBLP:journals/tit/AnarakiB17,pennington2015spherical}.  Denote $\C=[\C_1;\C_2;\cdots;\C_p]\in\R^{pq\times d}$, where $\C_t=\bT_t\A_t$ is compressed from $\A_t$ via $\bT_t=\S_t\H\D_t$.  Let $\l\le k=\min(m, d)$ and $\B\in \R^{\l\times d}$ is computed from FFD on $\C$. 
\end{definition}

We first derive the following lemma:  
\begin{Lemma}[FFD]\label{thm:FFD}
With the notations defined in Def.~\ref{def:theorem_notations} and with the probability at least $1-p\beta-(2p+1)\delta-\frac{2n}{e^{k}}$, we have
\begin{align}
&\|\A^T\A-\B^T\B\|_2\notag \\
&\leq O\Big(\frac{1}{\l}\log(\frac{md}{\beta})+\log(\frac{d}{\delta})\sqrt{\frac{k}{\l p^2}}+\sqrt{\log(\frac{d}{\delta})\frac{1+\sqrt{k/\l}}{p}}\Big)\|\A\|_F^2 \notag \\
&\leq\widetilde{O}\Big(\frac{1}{\l}+\Gamma(\l,p,k)\Big)\|\A\|_F^2,\label{eq:ffd_bound}
\end{align}
where $\Gamma(\l,p,k)=\sqrt{\frac{k}{\l p^2}}+\sqrt{\frac{1+\sqrt{k/\l}}{p}}$ with $p=\frac{n}{m}$ and $\widetilde{O}(\cdot)$ hides the logarithmic factors on $(\beta, \delta, k, d, m)$.

The time consumption of FFD is $\widetilde{O}(n\l^2\frac{d}{m}+nd)$ and its space requirement is $O(d\l)$.
\end{Lemma}
We defer the proof in Sec.~\ref{sec:ap2a}. 

\begin{remark}\label{re:my}
Comparing Lemma~\ref{thm:FFD} and the theoretical result of FD in~\cite{liberty2013simple}, we conclude the favorite characteristics of FFD:
\begin{compactitem}
\item The sketching error of FFD becomes smaller when $p$ increases, i.e., through increasing $n$ or decreasing $m$.  This is in line with our intuition because adding more training data will increase more information while  separating more blocks will enhance the sketching precision. Note that to make $\frac{n}{e^k}=\Theta(\frac{n}{e^d})\leq\Theta(1)$, we only need to have $d\geq\Theta(\log n)$, which satisfies in many practical cases. Then, we only require $p=\frac{n}{m}=\Theta(\frac{n}{d})\leq \Theta(\frac{n}{\log n})$, and the upper bound  $\Theta(\frac{n}{\log n})$ can be extremely large as $n$ increases. Thus,  $p$ can be extremely large to make $\Gamma(\l, p, k)=\sqrt{\frac{k}{\l p^2}}+\sqrt{\frac{1+\sqrt{k/\l}}{p}}$ negligible.
\item The time cost of FFD is inversely proportional to $m$.  A smaller $m$ will increase the time cost but enhance the sketching precision.  Hence, a proper $m$ is desirable.  
\item When $m$ is $\Theta(d)$, the time cost of FFD is $\widetilde{O}(n\l^2+nd)$.  {This is significantly superior to FD, $O(nd\l)$}, for the dense centered data when $\l\ll d$. {When more data come, i.e., $n$ increases, $p$ will become larger and the error bound of FFD tends to be tighter. }
\item In sum, FFD is more favorite to the big data applications when $1\ll d \ll n$ and it can attain the same estimation error bound as FD with lower computational cost.  
\end{compactitem}
\end{remark}

\begin{figure*}[htbp]
\centering
\subfigure{\includegraphics[width=.25\textwidth]{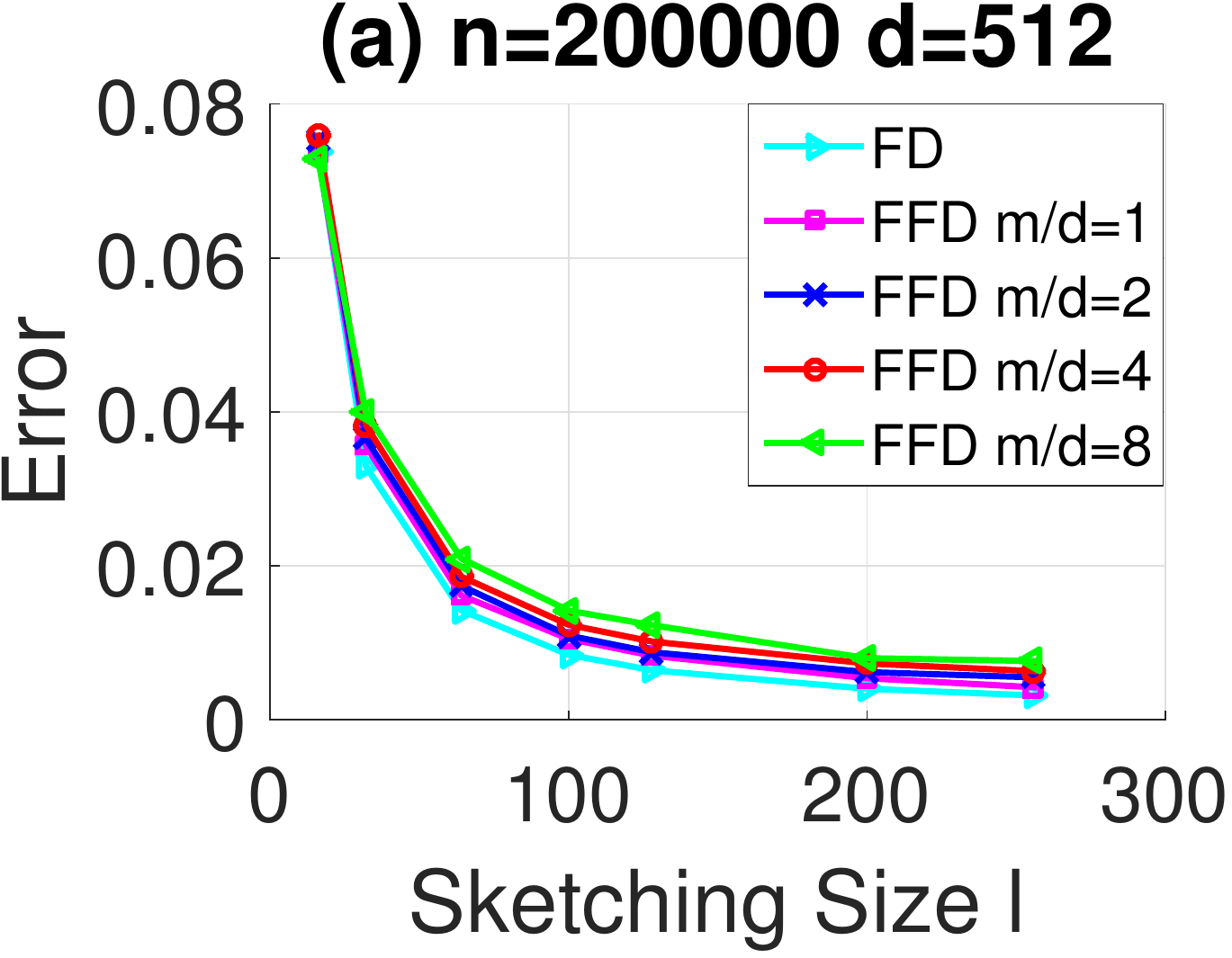}\label{fig:syn_acc_vl}}
\subfigure{\includegraphics[width=.25\textwidth]{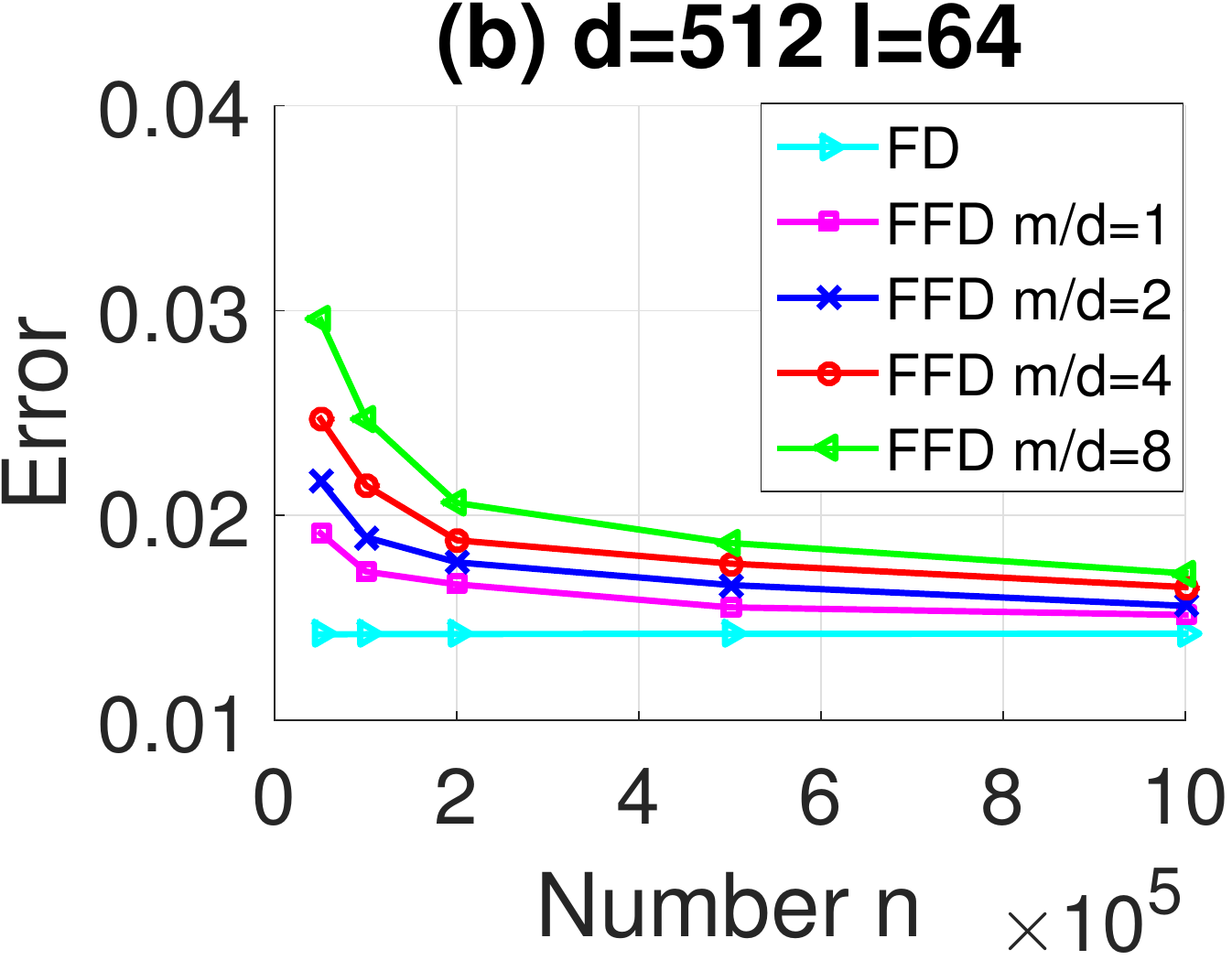}\label{fig:syn_acc_vn}}
\subfigure{\includegraphics[width=.25\textwidth]{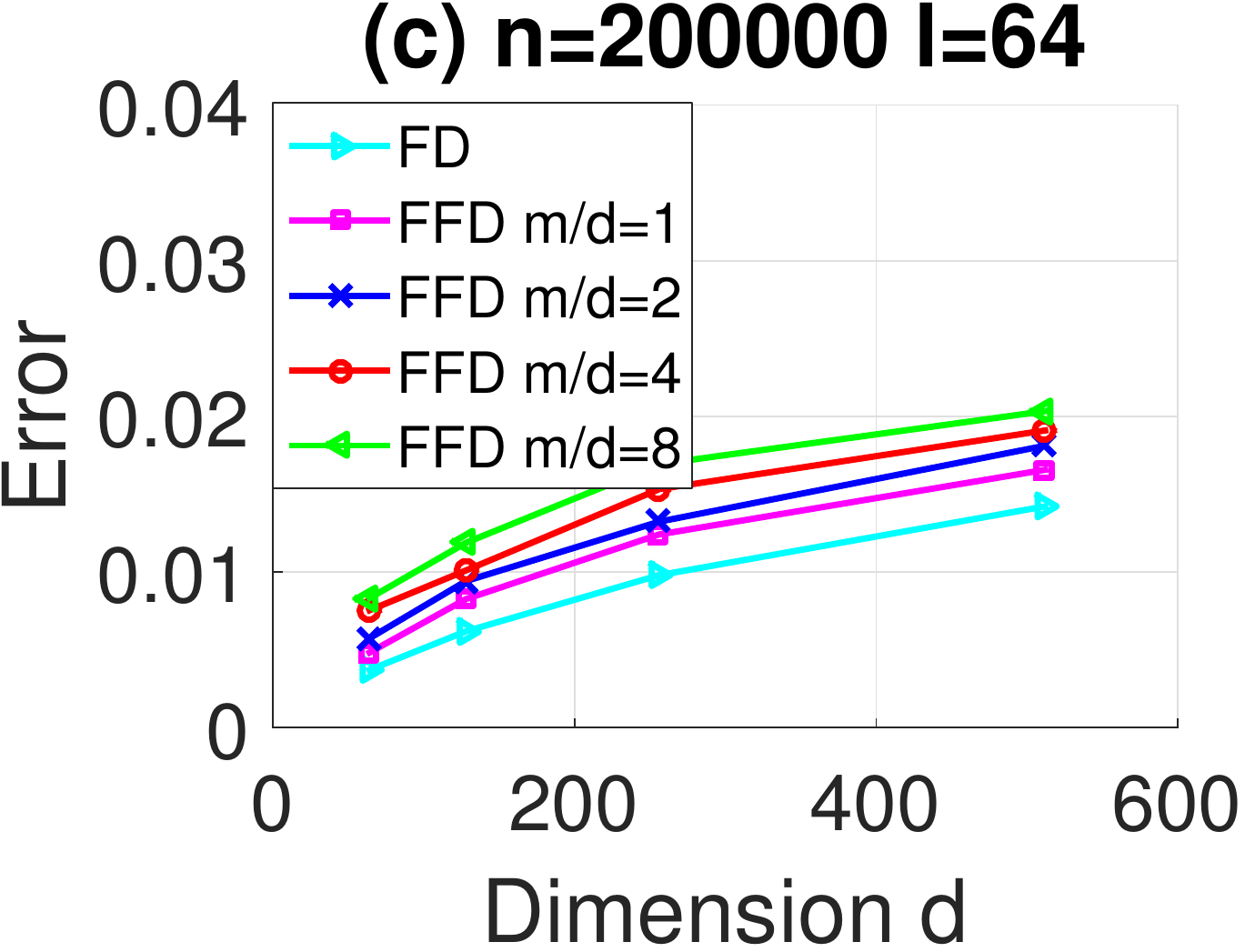}\label{fig:syn_acc_vd}}
\caption{Relative errors of FD and FFD under different settings.}
\label{fig:sketchingaccuracy}
\end{figure*}

\begin{figure*}[htbp]
\centering
\subfigure{\includegraphics[width=.25\textwidth]{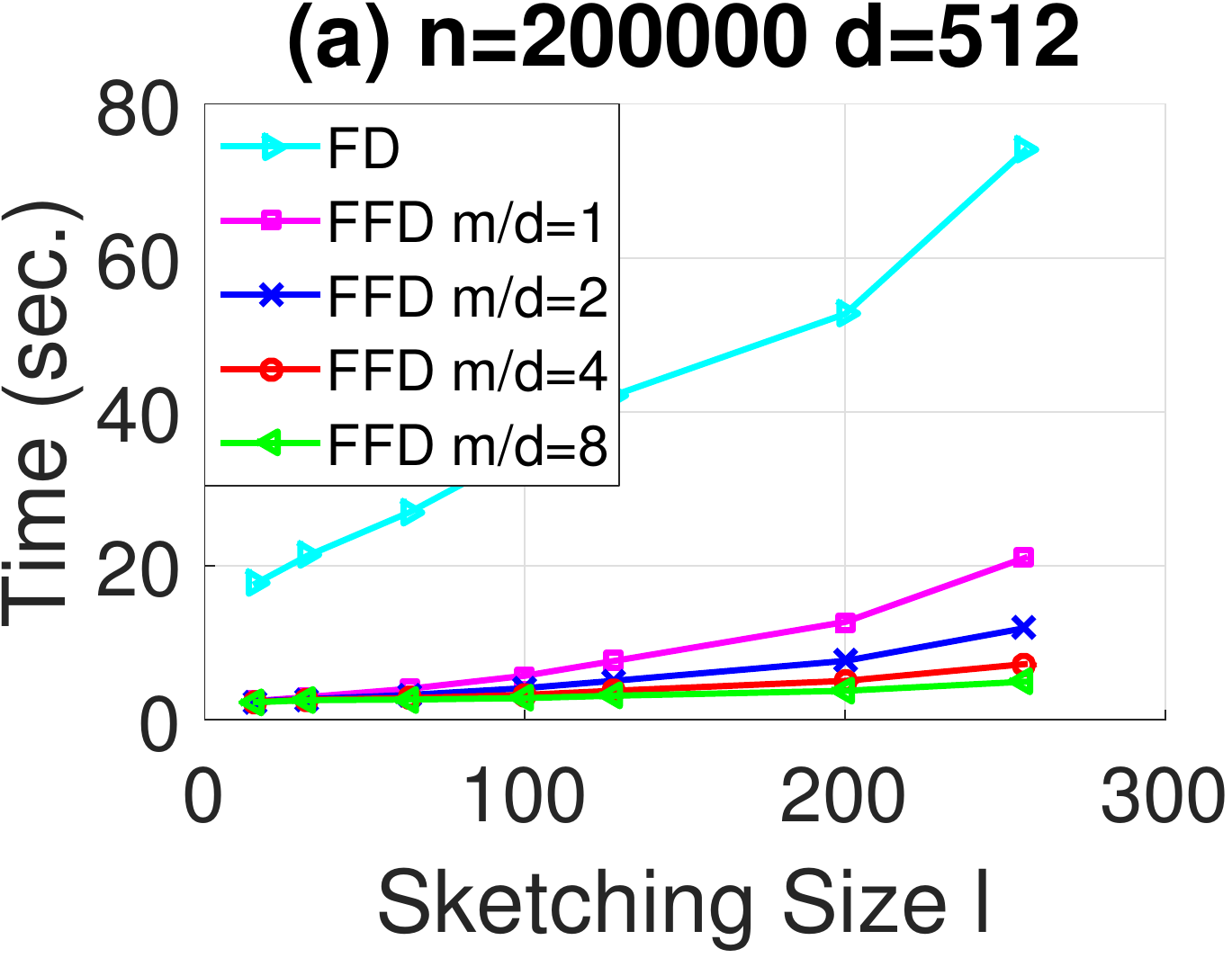}\label{fig:syn_time_vl}}
\subfigure{\includegraphics[width=.25\textwidth]{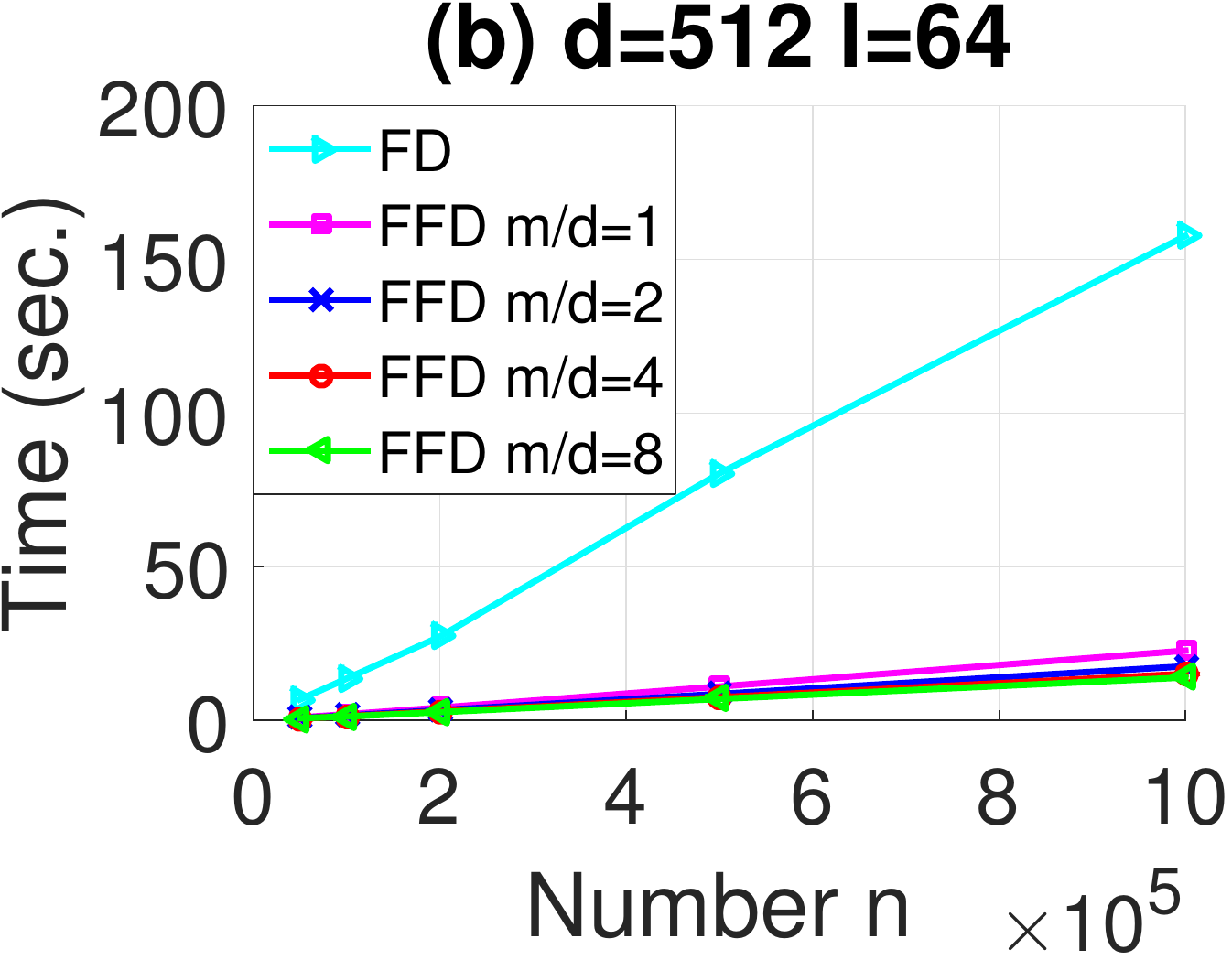}\label{fig:syn_time_vn}}
\subfigure{\includegraphics[width=.25\textwidth]{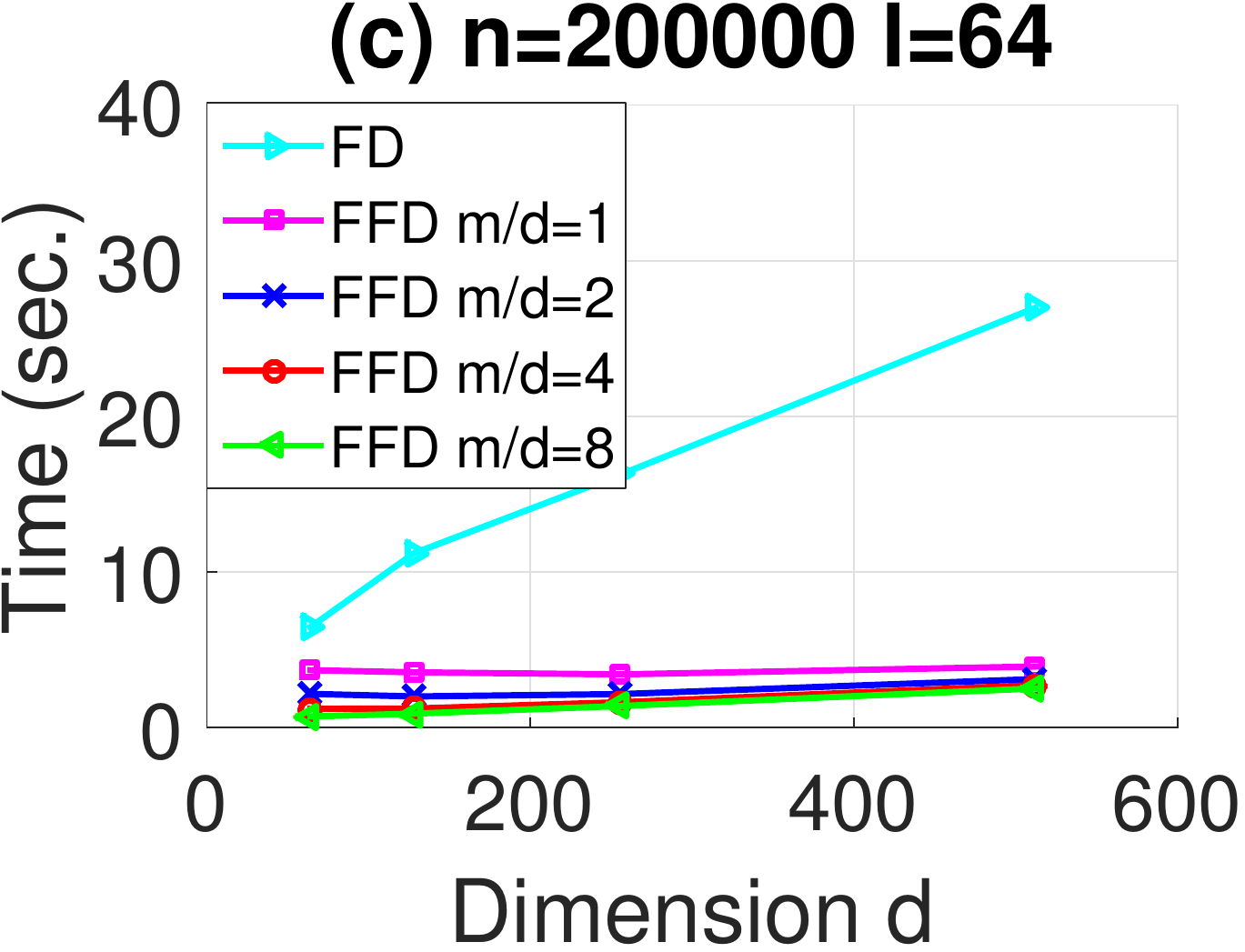}\label{fig:syn_time_vd}}
\caption{Time cost of FD and FFD under different settings.}
\label{fig:sketchingtime}
\end{figure*}
Based on Lemma~\ref{thm:FFD} and the online data centering mechanism, we derive the following theorem for FROSH: 
\begin{theorem}[FROSH]\label{thm:FROSH}
Given data $\A\in\R^{n\times d}$ with its row mean vector $\bmu\in\R^{1\times d}$, let the sketch $\B^{\l\times d}$ be constructed by FROSH in Algorithm~\ref{alg:frosh}. Then, with the probability defined in Lemma~\ref{thm:FFD}, we have
\begin{align}
\|(\A-\bmu)^T & (\A-\bmu)- \B^T\B\|_2 \notag \\\label{eq:frosh_bound}
&\leq\widetilde{O}\Big(\frac{1}{\l}+\Gamma(\l,p,k)\Big) \|\A-\bmu\|_F^2,
\end{align}
where $(\A-\bmu)\in\R^{n\times d}$ subtracts each row of $\A$ by $\bmu$, $\Gamma(\l,p,k)$ has been defined in Lemma~\ref{thm:FFD}, and $\W^T\in\R^{r\times d}$  is the hashing projection for Remark~\ref{rem:rotation} that  contains the top $r$ right singular vectors of $\B^{\l\times d}$.

The time cost of FROSH is $\widetilde{O}(n\l^2+nd+d\l^2)$ and the space cost is $O( d\l)$ when $m=\Theta(d)$ for invoking FFD of Algorithm~\ref{alg:ffd}.
\end{theorem}
The detailed proof is provided in Sec.~\ref{sec:ap2b}.  The primary analysis is similar to that of OSH~\cite{leng2015online}, where the learning accuracy can be maintained via accurate sketching.  Hence, we adopt the sketching error for the centered data $\A-\bmu$ to justify whether the sketching-based hashing achieves proper performance~\cite{leng2015online}.  We can observe from the bound in Eq.~(\ref{eq:frosh_bound}) that it significantly decreases from $O(nd\l)$ of OSH to $\widetilde{O}(n\l^2+nd)$ where usually in real-world applications, $1<\l\ll d\ll n$.  Note that, the relation between $\B$ and $\A-\bmu$ can theoretically lead to a similar relation between their projection matrices $\W_\B$ and $\W$, i.e., $\|(\A-\bmu)-(\A-\bmu)\W_\B\W_\B^T\|_2^2\leq (1+\epsilon)\|(\A-\bmu)-(\A-\bmu)\W\W^T\|_2^2$; see details in~\cite{DBLP:conf/uai/ChenKL17}.

\begin{theorem}[DFROSH]\label{thm:DFROSH}
Given the distributed data $\A=[\A_1;\A_2;\cdots;\A_\omega]\in\R^{n\times d}$ with its row mean vector $\bmu\in\R^{1\times d}$, let the sketch $\B^{\l\times d}$ be generated by DFROSH in Algorithm~\ref{alg:DFROSH}.  {Under the assumption that $n_i=n/\omega$ for $\A_i\in\R^{n_i\times d}$}, then with the probability and notations defined in Lemma~\ref{thm:FFD} we have
\begin{align}
\|(\A-\bmu)^T & (\A-\bmu)- \B^T\B\|_2 \notag \\\label{eq:dfrosh_bound}
&\leq\widetilde{O}\Big(\frac{1}{\l}+\Gamma(\l,p,k)\Big) \|\A-\bmu\|_F^2.
\end{align}
The time cost of DFROSH is the summation of  $\widetilde{O}(n\l^2/\omega+nd/\omega)$ in each distributed machine and  $O(\omega d\l^2)$ in the center machine.  {Totally, the time cost of DFROSH can be regarded as $\widetilde{O}(n\l^2/\omega+nd/\omega)$ when $\l\ll d\ll n$ and $\omega\leq O(\sqrt{n/d})$.}  In terms of the space cost, each machine consumes $O(d\l)$ space when $m=\Theta(d)$ for invoking FFD in Algorithm~\ref{alg:ffd}.
\end{theorem}
Hence, we can guarantee the sketching precision of DFROSH.  The detailed proof can be referred to Sec.~\ref{sec:ap2c}.

\section{Theoretical Proof}
\subsection{Preliminaries}\label{sec:preliminary}
Before providing the main theoretical results, we present four preliminary theoretical results for the rest proofs.  

The first theorem is the Matrix Bernstein inequality for the sum of independent zero-mean random matrices.
\begin{theorem}[{\cite{tropp2015introduction}}]\label{thm:bernstein}
Let $\{\A_i\}_{i=1}^L\in\R^{n\times d}$ be independent random matrices with  $\E\left[\A_i\right]=\mathbf{0}$ and $\|\A_i\|_2\leq R$ for all $i\in [L]$.  Let $\sigma^2=\max\{\|\sum_{i=1}^L\E\left[\A_i\A_i^T\right]\|_2, \|\sum_{i=1}^L\E\left[\A_i^T\A_i\right]\|_2\}$ be a variance parameter. Then, for all $\epsilon\geq 0$, we have
\begin{align}
\Pr\left(\left\|\sum_{i=1}^L\A_i\right\|_2\geq\epsilon\right)\leq (d+n)\exp\left(\frac{-\epsilon^2/2}{\sigma^2+R\epsilon /3}\right).
\end{align}
\end{theorem}

The second theorem characterizes the property of the compressed data via an SRHT matrix. 
\begin{theorem}[{\cite{lu2013faster}}]\label{thm:srhtnips} 
Given $\A\in\R^{m\times d}$, let $\mbox{rank}(\A)\leq k\leq \min(m,d)$ and $\bT\in\R^{q\times m}$ be an SRHT matrix. Then, with the probability at least $1-(\delta+\frac{m}{e^k})$ we have
\begin{align}
(1-\Delta)\A^T\A\preceq \A^T\bT^T\bT\A\ \preceq(1+\Delta)\A^T\A,
\end{align}
where $\Delta=\Theta(\sqrt{\frac{k\log(2k/\delta)}{q} })$.
\end{theorem}

By applying Corollary 3 in~\cite{DBLP:journals/tit/AnarakiB17}, one can directly derive the following lemma to bound the compressed data:
\begin{Lemma}\label{thm:srhtarxiv} 
Give data $\A\in\R^{m\times d}$, and an SRHT matrix $\bT\in\R^{q\times m}$. Then, with probability at least $1-\beta$,  we have 
 \begin{align}
 \|\bT\a_i\|_2&\leq \sqrt{2\log\left(\frac{2md}{\beta}\right)}\|\a_i\|_2, \, \forall  \, i\in [d] \label{eq:corollary3.4}.
 \end{align}
\end{Lemma}

Before proceeding, we also need the following Lemma to characterize the property of the scaled sampling matrix:
\begin{Lemma}\label{thm:ourlemma} 
Given data $\X\in\R^{m\times d}$, and a scaled sampling matrix $\S\in\R^{q\times m}$ in SRHT.  Then, we have
\begin{align}
\E[\X^T\S^T\S\X] &= \X^T\X.  \label{eq:i1} 
\end{align}
\end{Lemma}

\subsection{Proof of Lemma~\ref{thm:FFD}}\label{sec:ap2a}
\begin{proof}
We follow the notations defined in Def.~\ref{def:theorem_notations}. By the triangle inequality, we have
\begin{align}
\!\!\|\A^T\A-\B^T\B\|_2  \leq \|\A^T\A-\C^T\C\|_2+\|\C^T\C-\B^T\B\|_2.  \label{eq:final0}
\end{align}
Since $\B$ is computed from the standard FD on $\C$, then with the probability at least $1-p\beta$, we have
\begin{align}
\|\C^T\C-\B^T\B\|_2\leq \frac{2}{\l}\|\C\|_F^2 
\leq \frac{4}{\l}\log\left(\frac{2md}{\beta}\right)\|\A\|_F^2. \label{eq:th12}
\end{align}
The first inequality directly follows from that of FD~\cite{liberty2013simple}.  The second inequality holds by applying Lemma~\ref{thm:srhtarxiv} and the union bound on  $\|\C\|_F^2=\sum_{t=1}^p\|\C_t\|_F^2=\sum_{t=1}^p\sum_{i=1}^{d}\|\c_{t,i}\|_2^2$. 

Define $\X_t=\H\D_t\A_t$, we can start to bound $\|\A^T\A-\C^T\C\|_2$ by 
\begin{align}
\|\A^T\A&-\C^T\C\|_2=\left\|\sum_{t=1}^p(\A_t^T\A_t-\C_t^T\C_t)\right\|_2\notag \\&=\left\|\sum_{t=1}^p(\A_t^T\A_t-\X_t^T\S_t^T\S_t\X_t)\right\|_2. \label{eq:acbound}
\end{align}

Let $\Z_t=\A_t^T\A_t-\X_t^T\S_t^T\S_t\X_t$, $t\in[p]$, we obtain independent random variables, $\{\Z_t\}_{t=1}^p$.  By applying Lemma~\ref{thm:ourlemma}, we perform the expectation w.r.t. $\S_t$ and $\D_t$ and obtain 
\begin{align}
&\E[\X_t^T\S_t^T\S_t\X_t]=\E_{\D_t}\E_{\S_t}[\X_t^T\S_t^T\S_t\X_t|\D_t] \\
&=\E_{\D_t}[\X_t^T\X_t]=\E_{\D_t}[\A_t^T\D_t^T\H^T\H\D_t\A_t]=\A_t^T\A_t, \notag
\end{align}
where the second equality follows from Lemma~\ref{thm:ourlemma} by fixing $\D_t$ and applying the property of the unitary matrices on $\H$ and $\D_t$ to attain the last equality. Thus, $\{\Z_t\}_{t=1}^p$ satisfy the condition of the Matrix Bernstein inequality in Theorem~\ref{thm:bernstein}. 

By applying the union bound on Theorem~\ref{thm:srhtnips}, with the probability at least $1-(p\delta+\sum_{t=1}^p\frac{m}{e^{k_t}})$, we attain $\|\Z_t\|_2\leq \Delta_t \|\A_t^T\A_t\|_2=\Delta_t \|\A_t\|_2^2$ and  
\begin{align}
R=\max_{t\in [p]} \Delta_t \|\A_t\|_2^2, \label{eq:logterm2}
\end{align}
where $\Delta_t=\Theta(\sqrt{\frac{k_t\log(2k_t/\delta)}{q}})$ and $\mbox{rank}(\A_t)\leq k_t\leq \min(m,d)$.

{\bf Computing $\sigma^2$.} Due to the symmetry of each matrix $\Z_t$, we have $\sigma^2=\|\sum_{t=1}^p\E[(\Z_t)^2]\|_2$.  Hence, with the probability at least $1-(\delta+\frac{m}{e^{k_t}})$, we have \begin{align}
\mathbf{0}^{d\times d}&\preceq\E[(\Z_t)^2] \\
&=\E[(\X_t^T\S_t^T\S_t\X_t)^2]-(\A_t^T\A_t)^2 \label{eq:the213}\\
&\preceq\E[\|\S_t\X_t\|_2^2\X_t^T\S_t^T\S_t\X_t]-(\A_t^T\A_t)^2 \label{eq:the23}\\
&\preceq\E[(1+\Delta_t)\|\A_t\|_2^2\X_t^T\S_t^T\S_t\X_t]-(\A_t^T\A_t)^2 \label{eq:the233}\\
&=(1+\Delta_t)\|\A_t\|_2^2\A_t^T\A_t-(\A_t^T\A_t)^2 \label{eq:the25}
\end{align}
In the above, Eq.~(\ref{eq:the213}) and Eq.~(\ref{eq:the25}) hold  because $\E(\X_t^T\S_t^T\S_t\X_t)=\A_t^T\A_t$.  Eq.~(\ref{eq:the233}) follows from Theorem~\ref{thm:srhtnips}.  Eq.~(\ref{eq:the23}) holds because 
\begin{align}
&\mathbf{0}^{d\times d}\preceq(\X_t^T\S_t^T\S_t\X_t)^2\preceq\|\S_t\X_t\|_2^2\X_t^T\S_t^T\S_t\X_t,\notag
\end{align}
which results from the fact that for any $\y\in\R^d$,
\begin{align}
&\y^T(\X_t^T\S_t^T\S_t\X_t)^2\y=\|\y^T\X_t^T\S_t^T\S_t\X_t\|_2^2\notag\\
\leq& \|\y^T\X_t^T\S_t^T\|_2^2\|\S_t\X_t\|_2^2=\|\S_t\X_t\|_2^2\y^T\X_t^T\S_t^T\S_t\X_t\y. \notag
\end{align}
Then, we have
\begin{align}
&\left\|\sum_{t=1}^p\E[(\Z_t)^2]\right\|_2\leq \sum_{t=1}^p\|\E[(\Z_t)^2]\|_2\notag\\
\leq& \sum_{t=1}^p\Big\|(1+\Delta_t)\|\A_t\|_2^2\A_t^T\A_t-(\A_t^T\A_t)^2\Big\|_2 \label{eq:the3311}\\
=&\sum_{t=1}^p\Big\|(1+\Delta_t)\|\A_t\|_2^2\U_t\bS_t^2\U_t-\U_t\bS_t^4\U_t\Big\|_2\label{eq:the32}\\
=&\sum_{t=1}^p\Big\|(1+\Delta_t)\|\A_t\|_2^2\bS_t^2-\bS_t^4\Big\|_2\notag\\
=&\sum_{t=1}^p\max_{j\in[d]}|(1+\Delta_t)\sigma_{t1}^2\sigma_{tj}^2-\sigma_{tj}^4|\notag\\
\leq&\sum_{t=1}^p(1+\Delta_t)\sigma_{t1}^4=\sum_{t=1}^p(1+\Delta_t)\|\A_t\|_2^4\notag\\
\leq&\max_{t\in [p]}p(1+\Delta_t)\|\A_t\|_2^4.\notag
\end{align}
where Eq.~(\ref{eq:the3311}) holds due to Eq.~(\ref{eq:the25}) and $\U_t$ in Eq.~(\ref{eq:the32}) is computed from the SVD of $\A_t$, i.e., $\A_t=\U_t\bS_t\V_t^T$ and the eigenvalues $\sigma_{tj}\triangleq \sigma_{t,jj}$ listed in the descending order in $\bS_t$.  

By Theorem~\ref{thm:bernstein}, we have
\begin{align}
\Pr(\left\|\sum_{t=1}^p\Z_t\right\|_2\geq\epsilon)\leq 2d\exp(\frac{-\epsilon^2/2}{\sigma^2+R\epsilon /3}). \label{eq:thfin}
\end{align}

Let $\delta$ denote the RHS of Eq.~(\ref{eq:thfin}), we can get  \begin{align}
\epsilon&=\log\left(\frac{2d}{\delta}\right)\left(\frac{R}{3}+\sqrt{\left(\frac{R}{3}\right)^2+\frac{2\sigma^2}{\log(2d/\delta)}}\right) \label{eq:logterm1} \\
\leq& \log\left(\frac{2d}{\delta}\right)\frac{2R}{3}+\sqrt{2\sigma^2\log\left(\frac{2d}{\delta}\right)}\\
\leq&\max_{t\in [p]}\widetilde{O}\left(\Delta_t\|\A_t\|_2^2\right)+\max_{t\in [p]}\widetilde{O}\left(\sqrt{p(1+\Delta_t)}\|\A_t\|_2^2\right) \label{eq:final12}\\
\leq& \max_{t\in [p]}\widetilde{O}\left(\left(\sqrt{\frac{k}{\l p^2}}+\sqrt{\frac{1+\sqrt{k/\l}}{p}}\right)\frac{\|\A_t\|_2^2}{\|\A_t\|_F^2}\right)\|\A\|_F^2 \label{eq:final2}\\
\leq &\widetilde{O}\left(\left(\sqrt{\frac{k}{\l p^2}}+\sqrt{\frac{1+\sqrt{k/\l}}{p}}\right)\right)\|\A\|_F^2.  \label{eq:finalfinal2}
\end{align}

To derive Eq.~(\ref{eq:final2}) from Eq.~(\ref{eq:final12}), we first substitute $\Delta_t=\Theta\left(\sqrt{\frac{k_t\log(2k_t/\delta)}{q}}\right)$ into Eq.~(\ref{eq:final12}) and set $k=k_t=\min(m,d)$, which allows Eq.~(\ref{eq:final12}) to become the maximum of the sum of two functions. Then, we take the definition $q=\l/2$ and apply a common practical assumption of that $p\lambda_1 \leq \|\A\|_F^2=\sum_{t=1}^p\|\A_t\|_F^2\leq p\lambda_2$ with each $\|\A_t\|_F^2$ bounded between $\lambda_1$ and $\lambda_2$ that are very close to each other.

Combing Eq.~(\ref{eq:finalfinal2}) with Eq.~(\ref{eq:final0}) and Eq.~(\ref{eq:th12}) based on the union bound, we obtain the desired result with the probability at least $1-p\beta-(2p+1)\delta-2p\frac{m}{e^{k}}$.

The computational analysis is straightforward based on that in FD and~Sec.\ref{subsection:imple}. 
\end{proof}

\subsection{Proof of Theorem~\ref{thm:FROSH}}\label{sec:ap2b}
\begin{proof}
By applying Lemma~\ref{thm:FFD} and the proof of OSH~\cite{leng2015online},  we can derive the theoretical error bound.


\end{proof}

\subsection{Proof of Theorem~\ref{thm:DFROSH}}\label{sec:ap2c}

\begin{proof}
The proof resembles those in Lemma~\ref{thm:FFD} and Theorem~\ref{thm:FROSH}, where we can first bound the sketching without considering the \textit{online centering procedure}. Given all distributed data $\A$, we have 
\begin{align}
\|\A^T\A-\B^T\B\|_2 \leq \|\A^T\A-\C^T\C\|_2+\|\C^T\C-\B^T\B\|_2 \notag,
\end{align}
where $\C$ is compressed from $\A$ by the fast transformations involved in FROSH in step~\ref{step:disosh:dis} of Algorithm~\ref{alg:DFROSH}, and $\B$ is the finally sketched data for $\C$  as shown in steps~\ref{step:bigB1} and~\ref{step:disosh:s21} of Algorithm~\ref{alg:DFROSH}.

First, we bound $\|\C^T\C-\B^T\B\|_2$. Without loss of generality, we compress $\C_i$ from the distributed input $\A_i$ by the fast transformations involved in FROSH in the  step ~\ref{step:disosh:dis} of Algorithm~\ref{alg:DFROSH} and obtain the sketch $\B_i$ for $\C_i$ via FD, i.e., $\B_i=\text{FD}(\C_i)$, where $i=1, \ldots, \omega$. Therefore, we obtain  
\begin{align}
&\|\C^T\C-\B^T\B\|_2 \notag
\\=& \|\C^T\C-\text{FD}([\B_1;\B_2;\cdots;\B_\omega])^T\text{FD}([\B_1;\B_2;\cdots;\B_\omega])\|_2  \notag
\\\leq& \frac{2}{\l}\|\C\|_F^2 \label{eq:d:1_2}
\\\leq& \frac{4}{\l}\log\left(\frac{2md}{\beta}\right)\|\A\|_F^2, \label{eq:d:1_3}
\end{align}
{where Eq.~(\ref{eq:d:1_2}) holds due to the distributed or parallelizable property of FD as proved in Section 2.2 of~\cite{liberty2013simple}, and Eq.~(\ref{eq:d:1_3}) follows Eq.~(\ref{eq:th12}).}



Next, we bound $\|\A^T\A-\C^T\C\|_2$ in DFROSH.  Without loss of generality, we let the distributed input $\A=\{\A_i\}_{i=1}^\omega$, or $\A=[\A_{11};\cdots;\A_{1p_1};\A_{21};\cdots;\A_{\omega p_\omega}]\in\R^{n\times d}$, where $\A_{it}\in\R^{m\times d}$ is the independent small data chunks that are processed by each iteration of FROSH involved in step 2 of  Algorithm~\ref{alg:DFROSH}, $p_i=\frac{n_i}{m}$, $i=1, \ldots, \omega$, and $t=1, \ldots, p_\omega$. Then, we have
\begin{align}
&\|\A^T\A-\C^T\C\|_2=\left\|\sum_{i=1}^\omega\A_i^T\A_i-\C_i^T\C_i\right\|_2\notag
\\=&\left\|\sum_{i=1}^\omega\sum_{t=1}^{p_i}\A_{it}^T\A_{it}-\C_{it}^T\C_{it}\right\|_2=\left\|\sum_{j=1}^p\A^{jT}\A^j-\C^{jT}\C^j\right\|_2 \notag
\\ \leq&\widetilde{O}\Big(\frac{1}{\l}+\Gamma(\l,p,k)\Big) \|\A\|_F^2, \label{eq:d:2_3}
\end{align}
where we denote $\A_{it}$ and $\C_{it}$ to $\A^j$ and $\C^j$, respectively, because $\sum_{i=1}^\omega p_i=p$.  Following the proof of Eq.~(\ref{eq:finalfinal2}), we then yield  Eq.~(\ref{eq:d:2_3}). 

Finally, {we combine Eq.~(\ref{eq:d:1_3}) and Eq.~(\ref{eq:d:2_3}) to derive the corresponding bound and conclude the whole proof.} 
\end{proof}

\section{Experiments}
In the experiments, we address the following issues:
\begin{compactenum}
\item What are different manifestations of FD and FFD in terms of sketching precision and time cost?
\item What is the performance of FROSH and DFROSH comparing with other online algorithms and batch-trained algorithms?
\end{compactenum}
We conduct experiments on synthetic datasets to answer the first question and real-world datasets to answer the second question, respectively.  For fair comparisons, all the experiments are conducted in MATLAB R2015a in the mode of a single thread running on standard workstations {with Intel CPU@2.90GHz, $128$GB RAM and the operating system of Linux.}  All the results are averaged over 10 independent runs. 

\begin{figure*}[htbp]
\setlength{\parskip}{-0.23cm}
\setlength{\abovecaptionskip}{0.13cm}
\setlength{\belowcaptionskip}{-0.13cm}
\centering
\subfigure{\includegraphics[width=.25\textwidth]{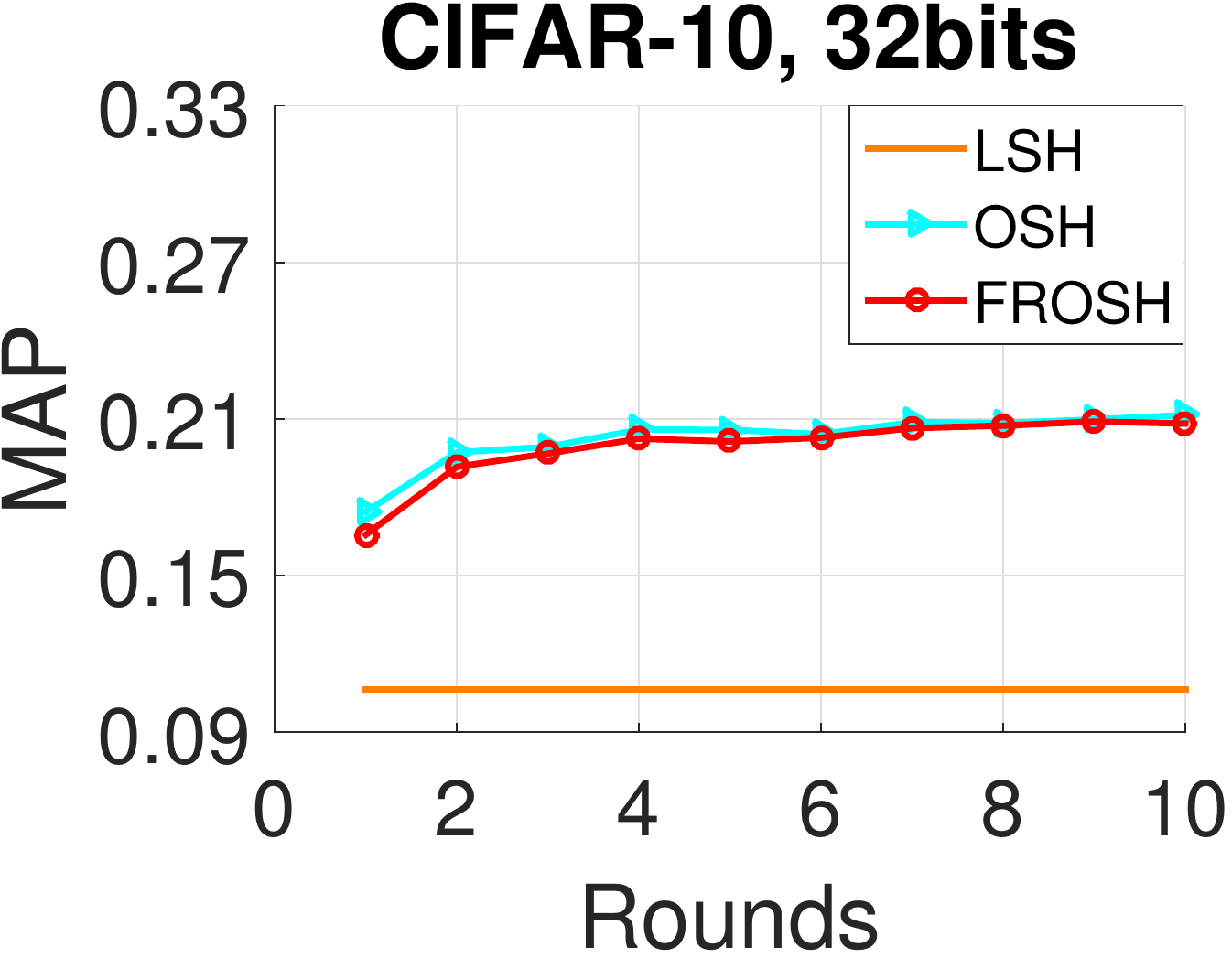}\label{fig:acc_cifar10_32}}
\subfigure{\includegraphics[width=.25\textwidth]{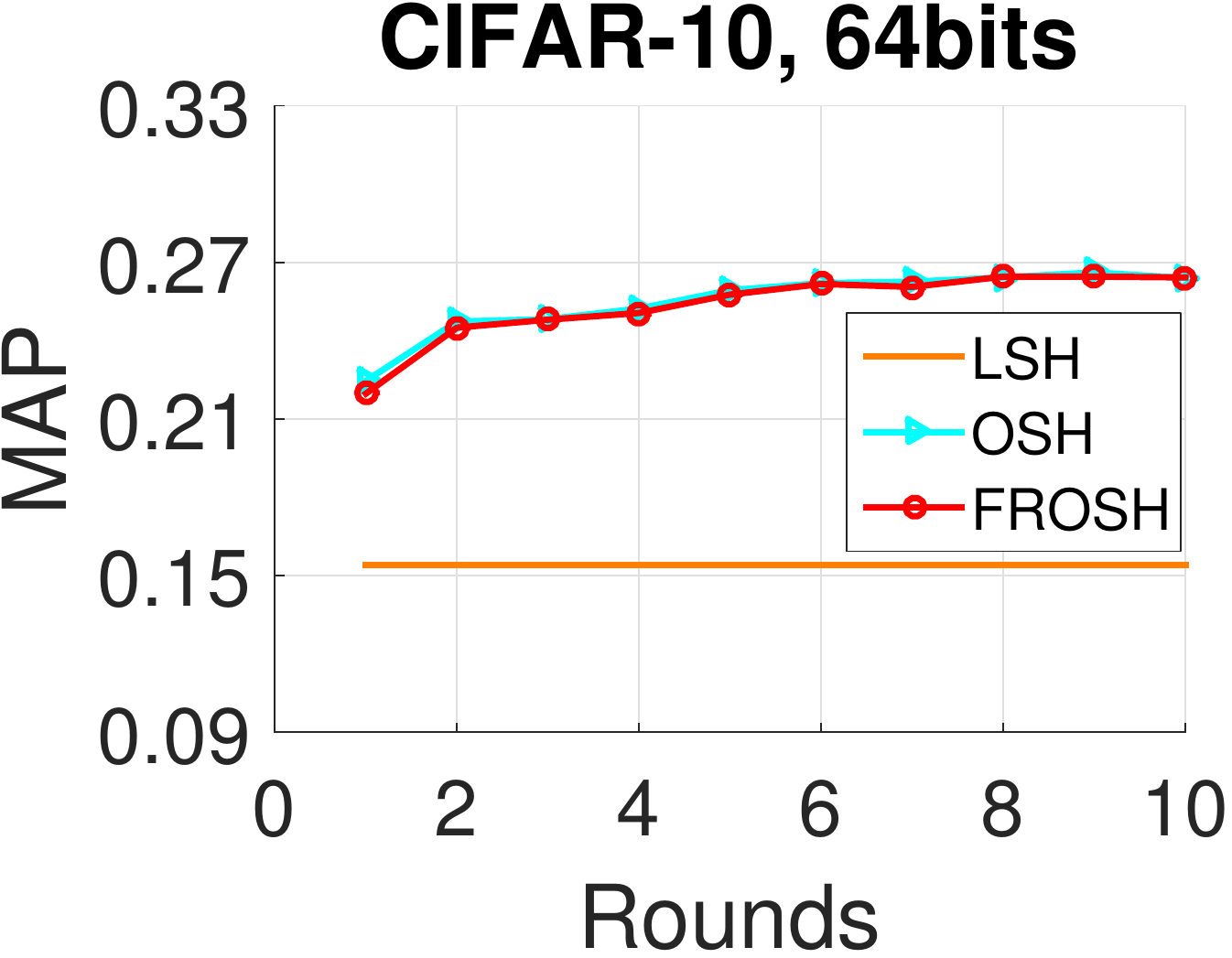}\label{fig:acc_cifar10_64}}
\subfigure{\includegraphics[width=.25\textwidth]{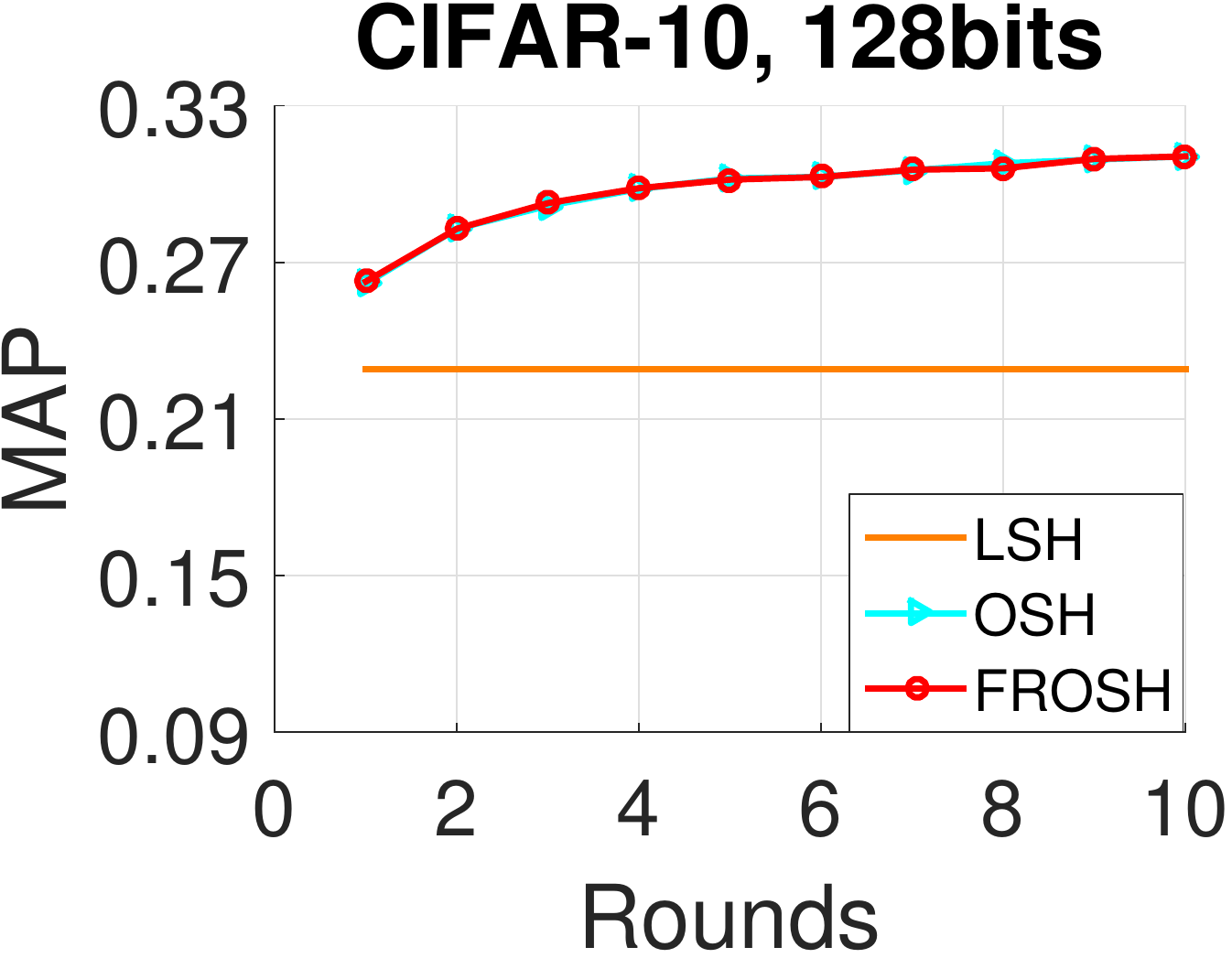}\label{fig:acc_cifar10_128}}
\subfigure{\includegraphics[width=.25\textwidth]{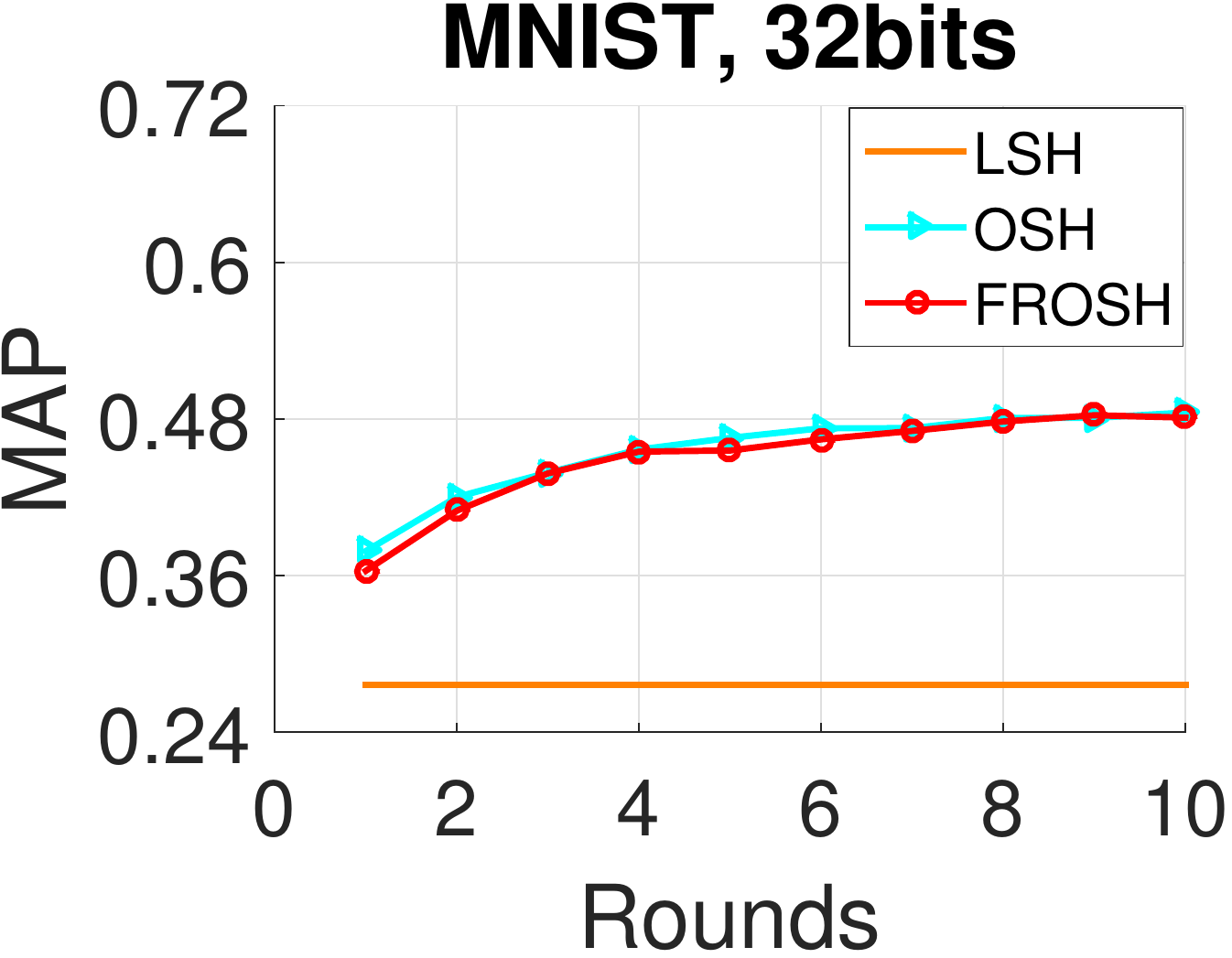}\label{fig:acc_mnist_32}}
\subfigure{\includegraphics[width=.25\textwidth]{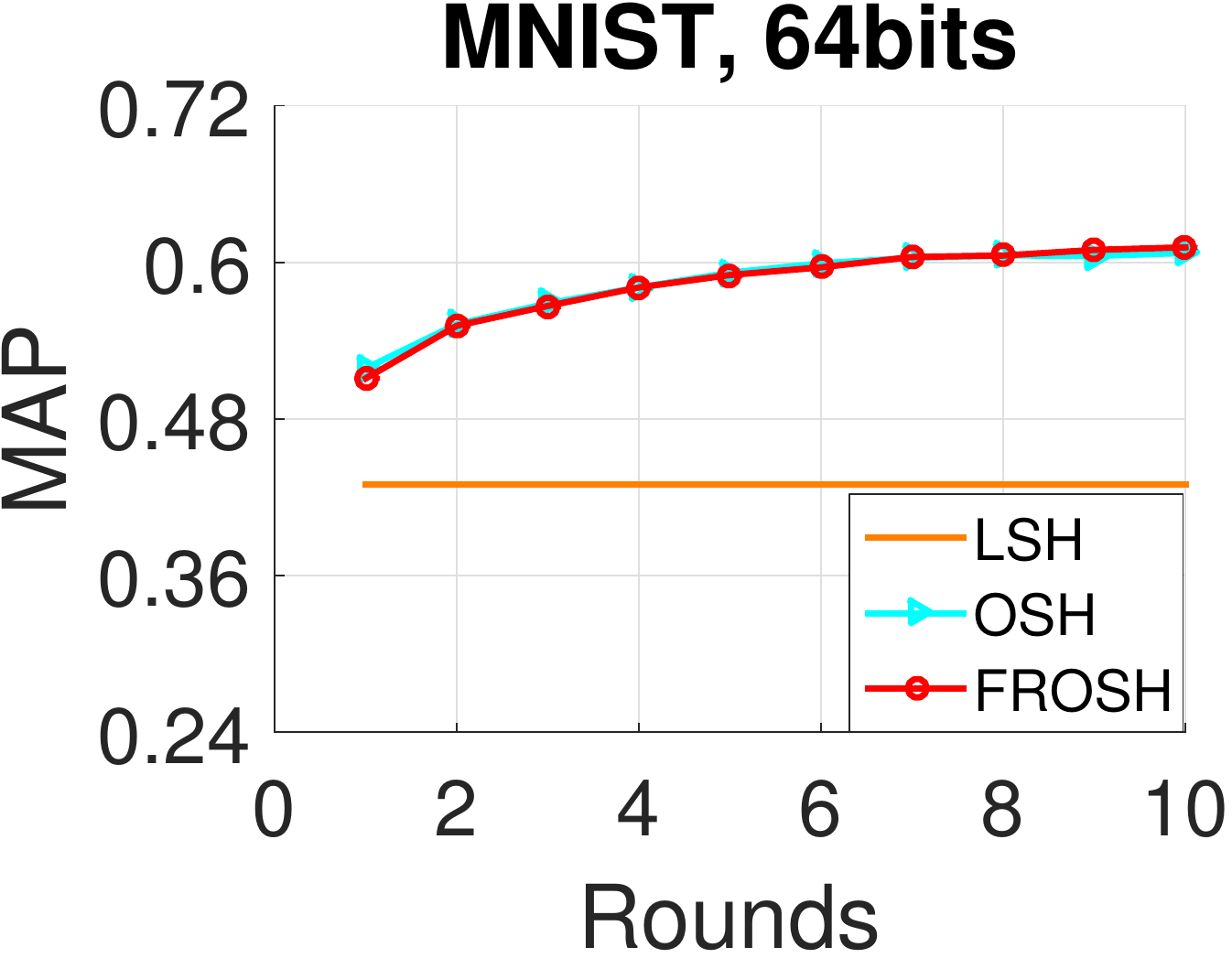}\label{fig:acc_mnist_64}}
\subfigure{\includegraphics[width=.25\textwidth]{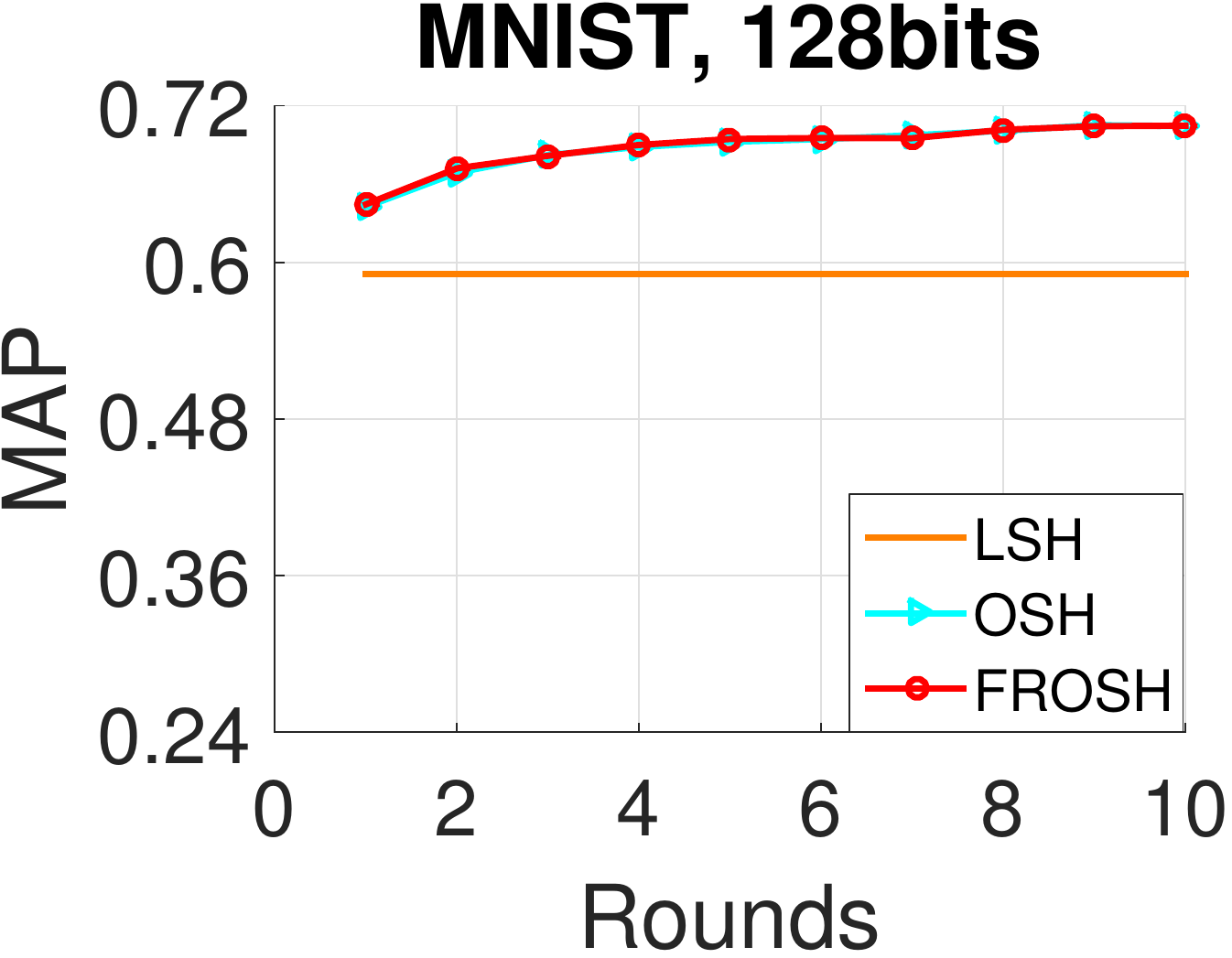}\label{fig:acc_mnist_128}}
\subfigure{\includegraphics[width=.25\textwidth]{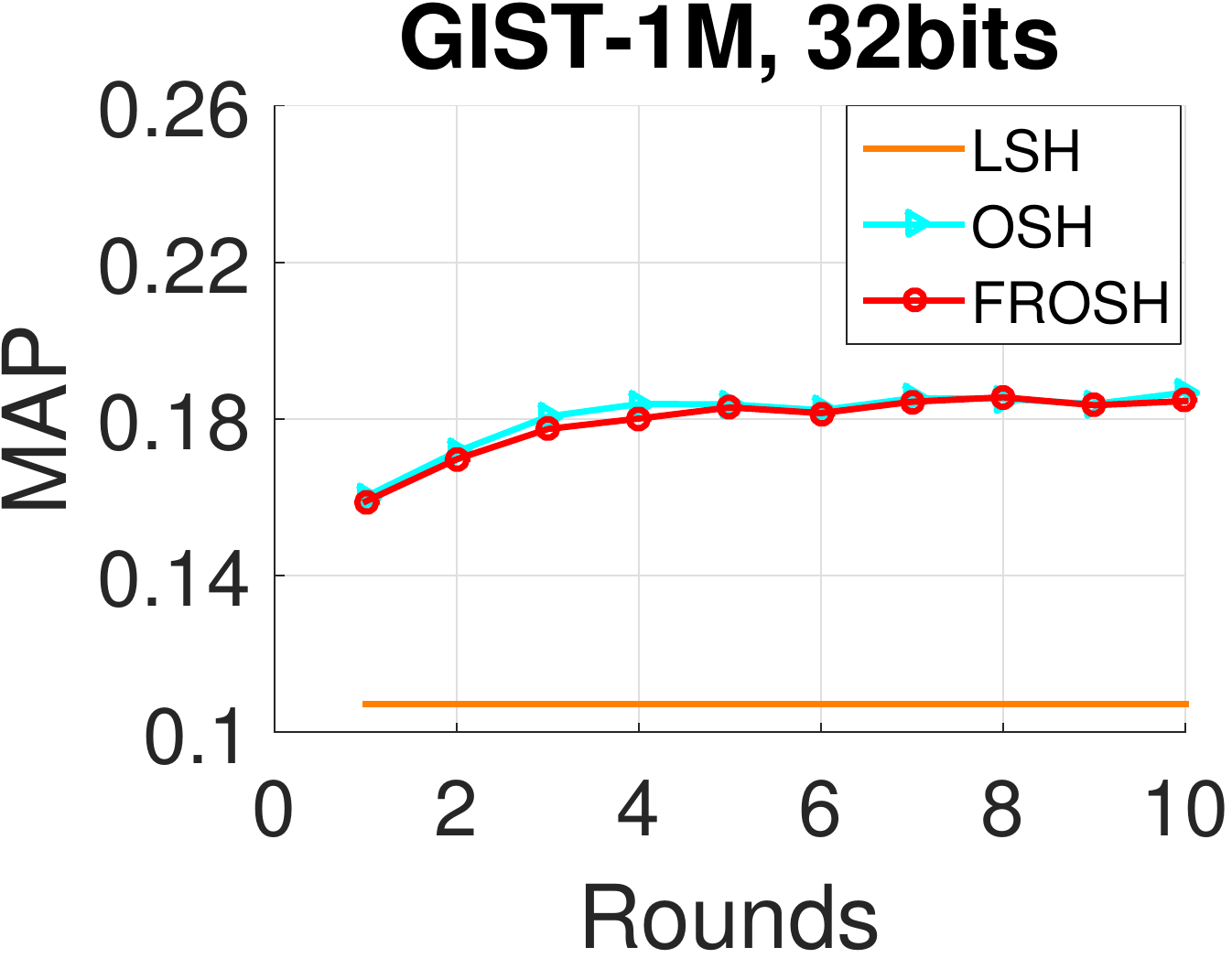}\label{fig:acc_gist_32}}
\subfigure{\includegraphics[width=.25\textwidth]{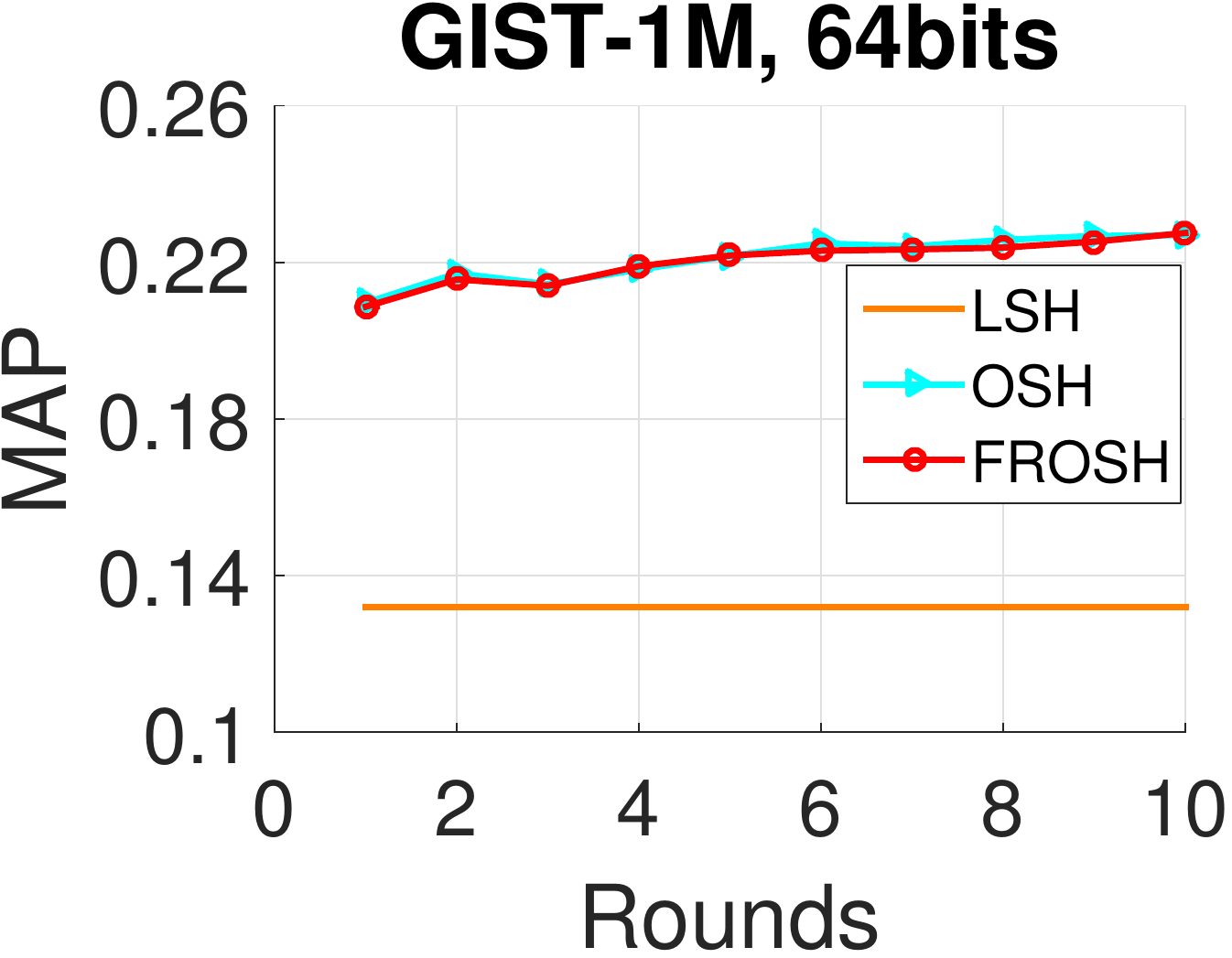}\label{fig:acc_gist_64}}
\subfigure{\includegraphics[width=.25\textwidth]{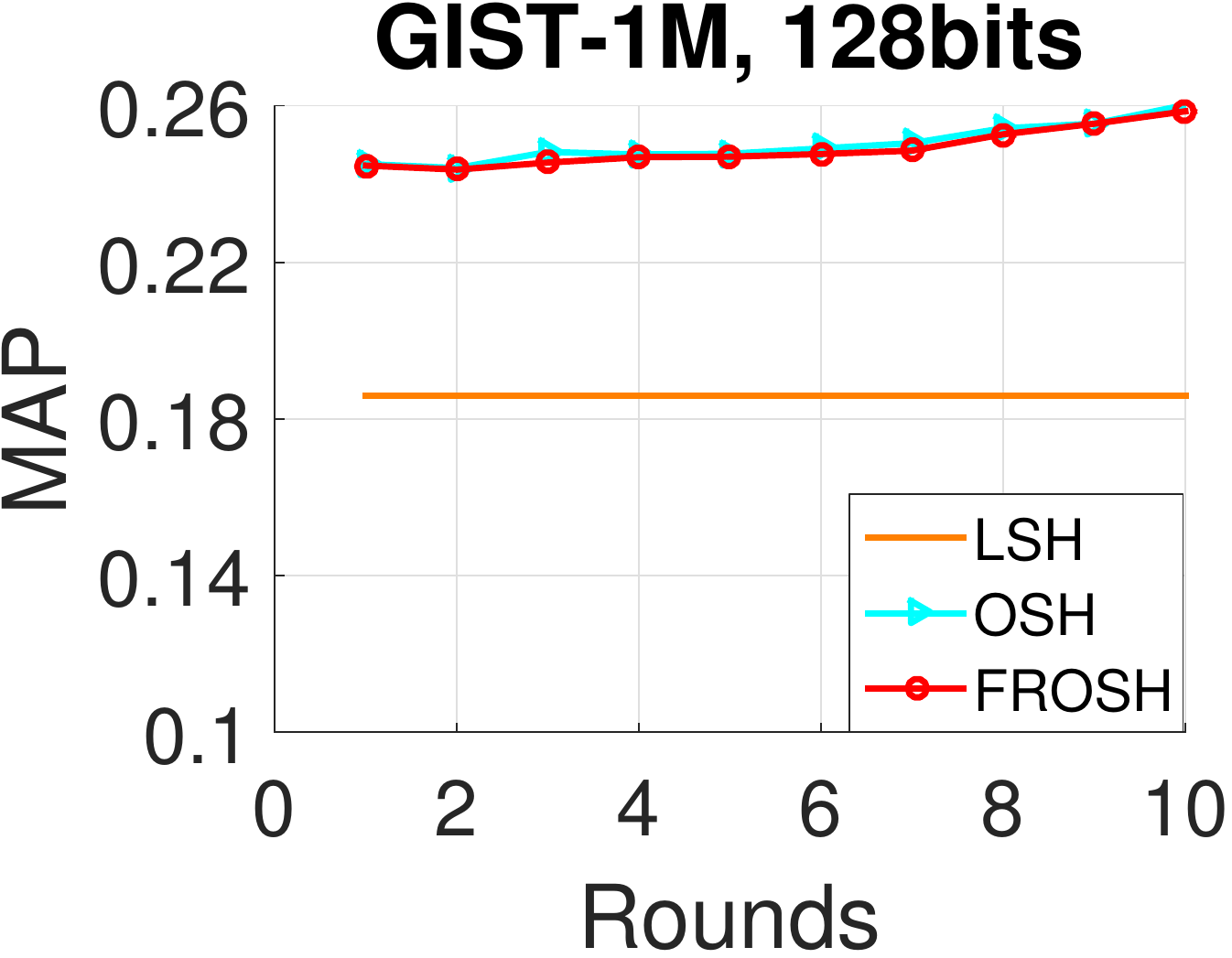}\label{fig:acc_gist_128}}
\subfigure{\includegraphics[width=.25\textwidth]{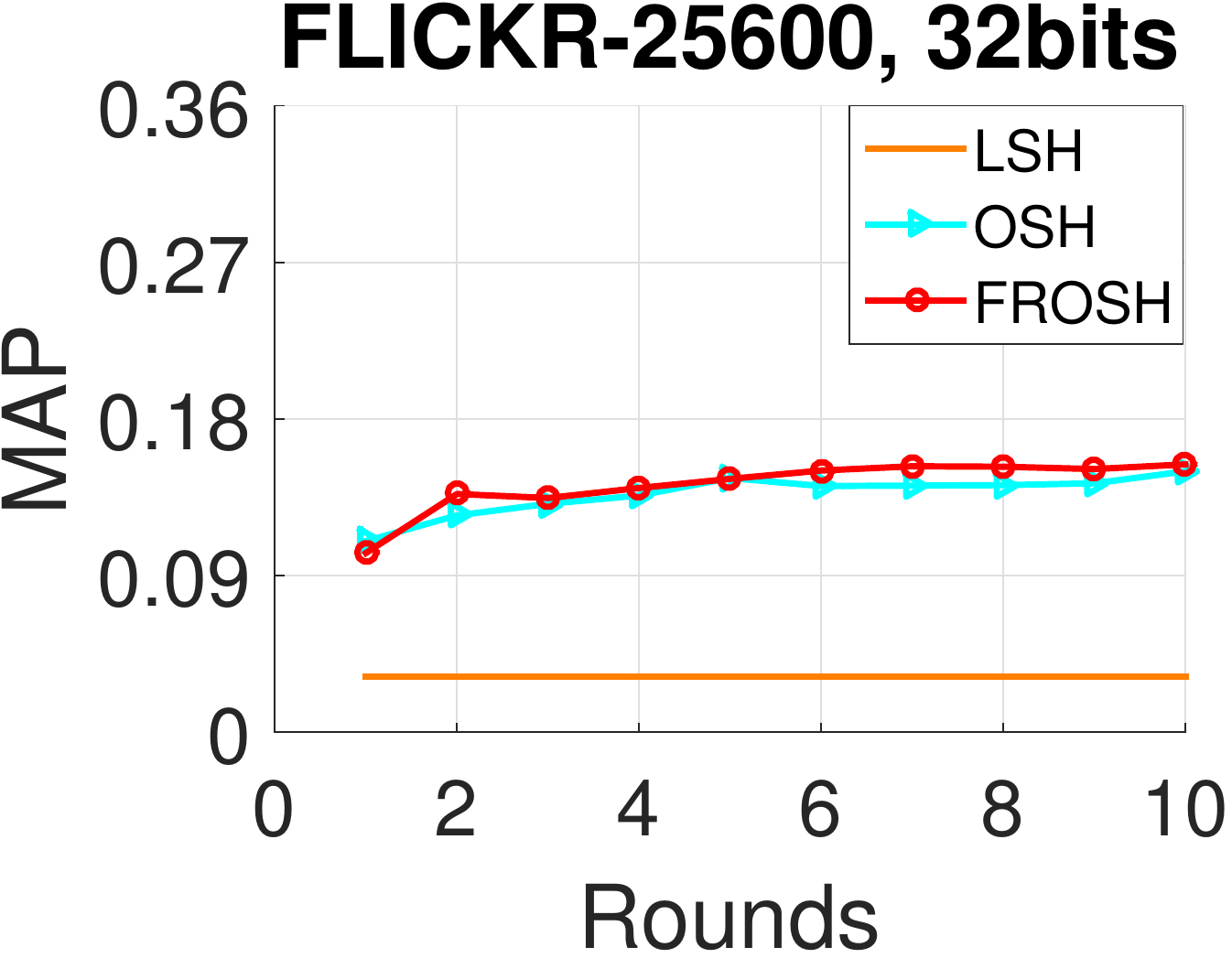}\label{fig:acc_flicker_32}}
\subfigure{\includegraphics[width=.25\textwidth]{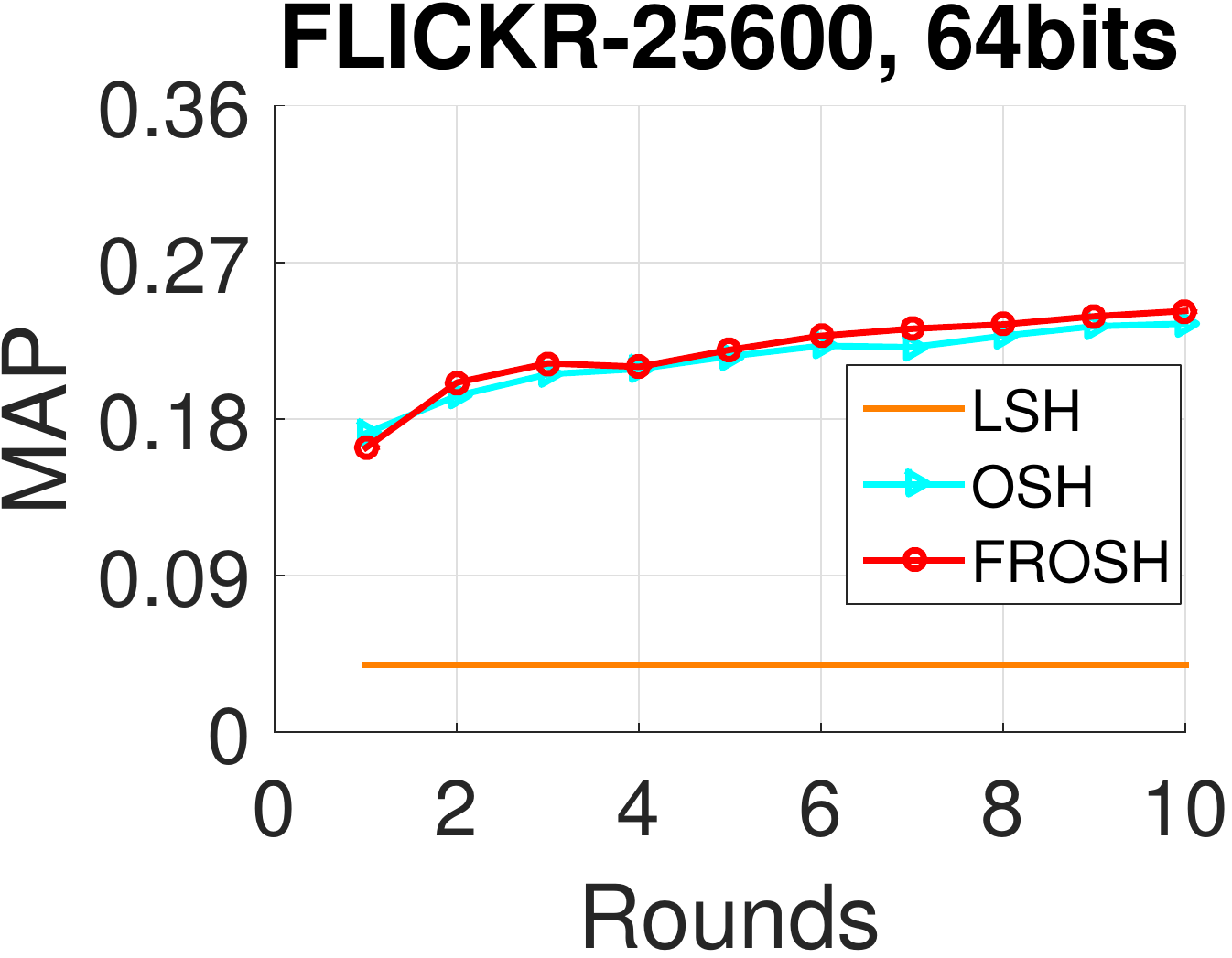}\label{fig:acc_flicker_64}}
\subfigure{\includegraphics[width=.25\textwidth]{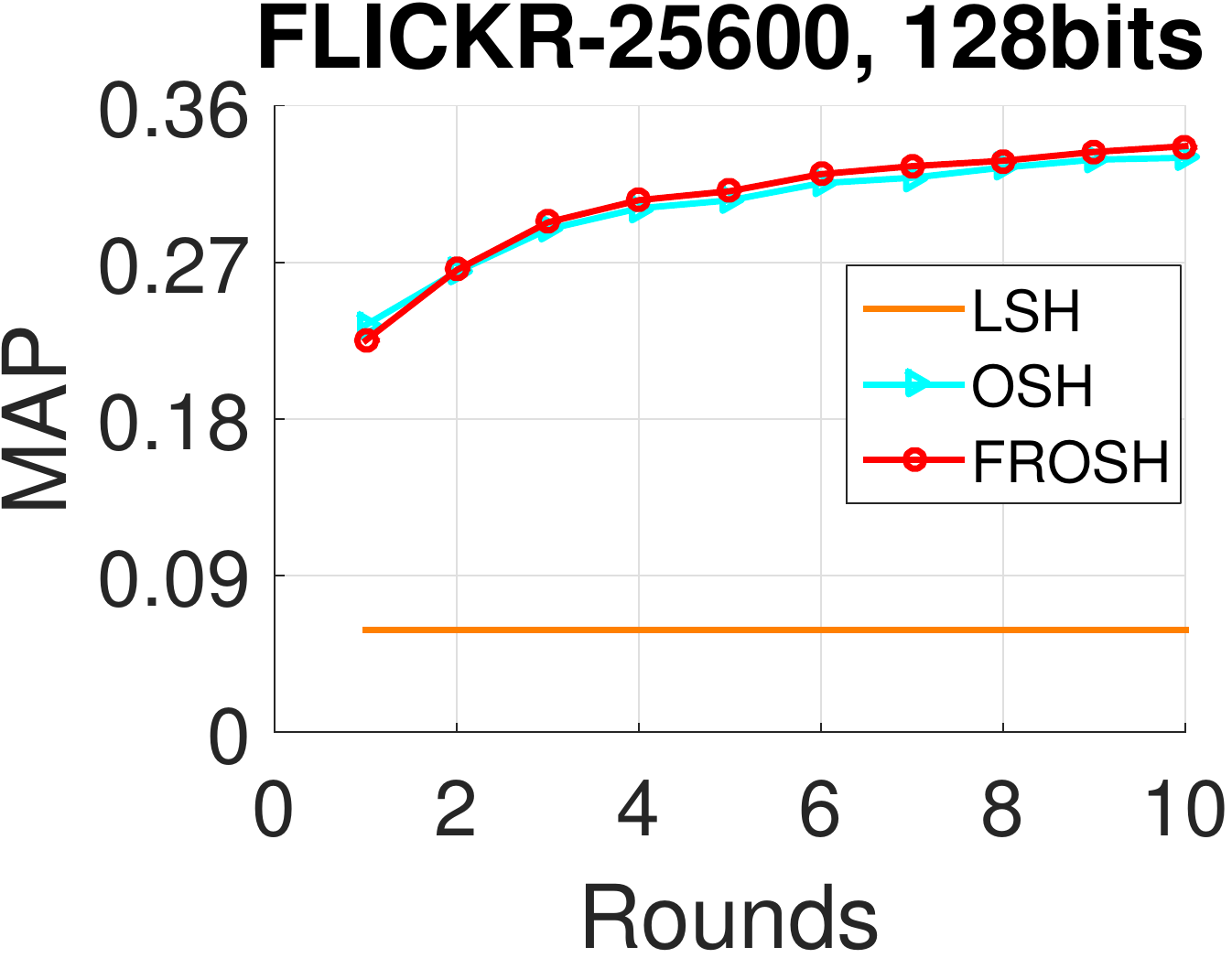}\label{fig:acc_flicker_128}}
\caption{MAP at each round with 32, 64, and 128 bits for online hashing: LSH, OSH, and FROSH.}
\label{fig:map}
\end{figure*}

\begin{figure*}[htbp]
\centering
\subfigure
{\includegraphics[width=.24\textwidth]{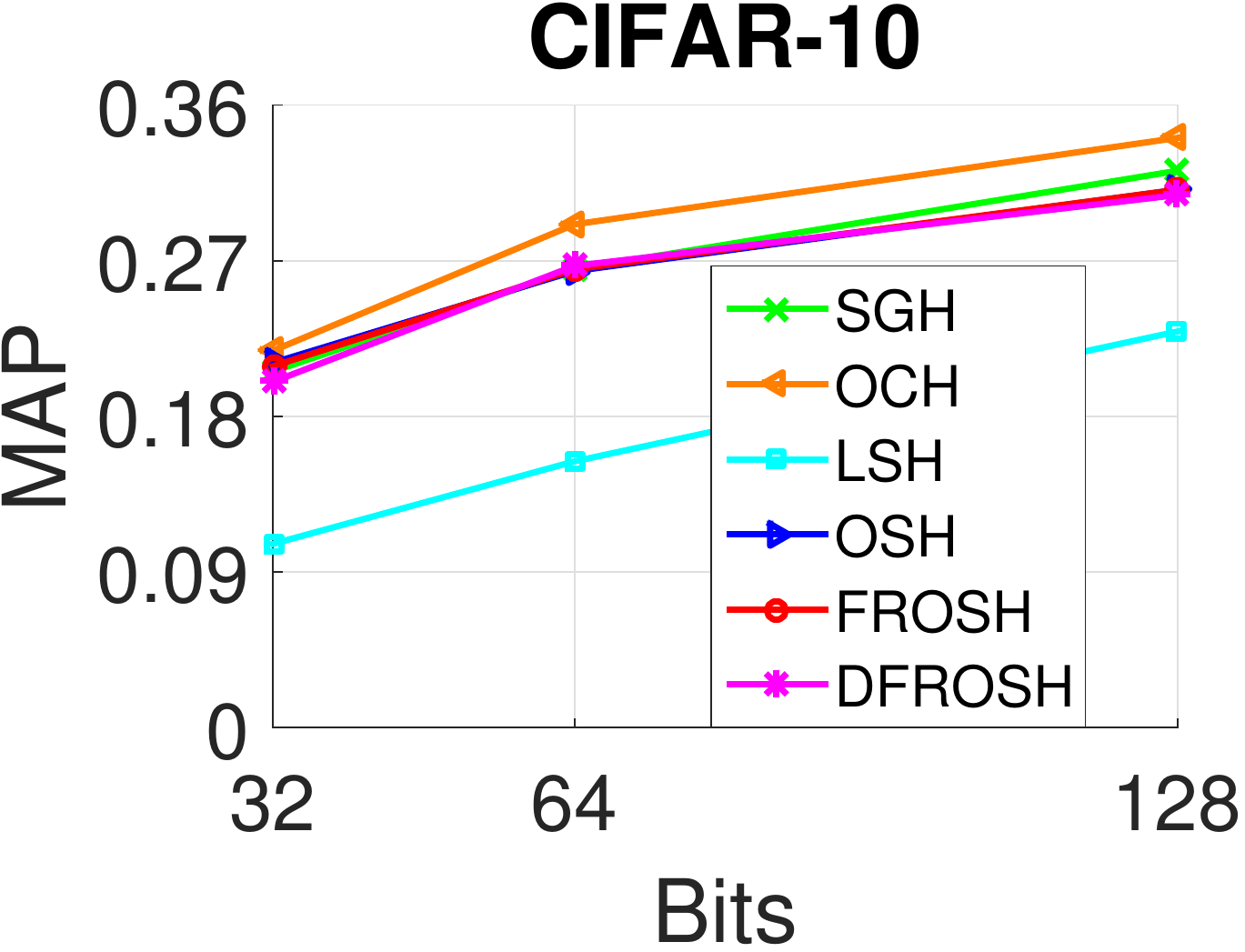}}
\subfigure
{\includegraphics[width=.24\textwidth]{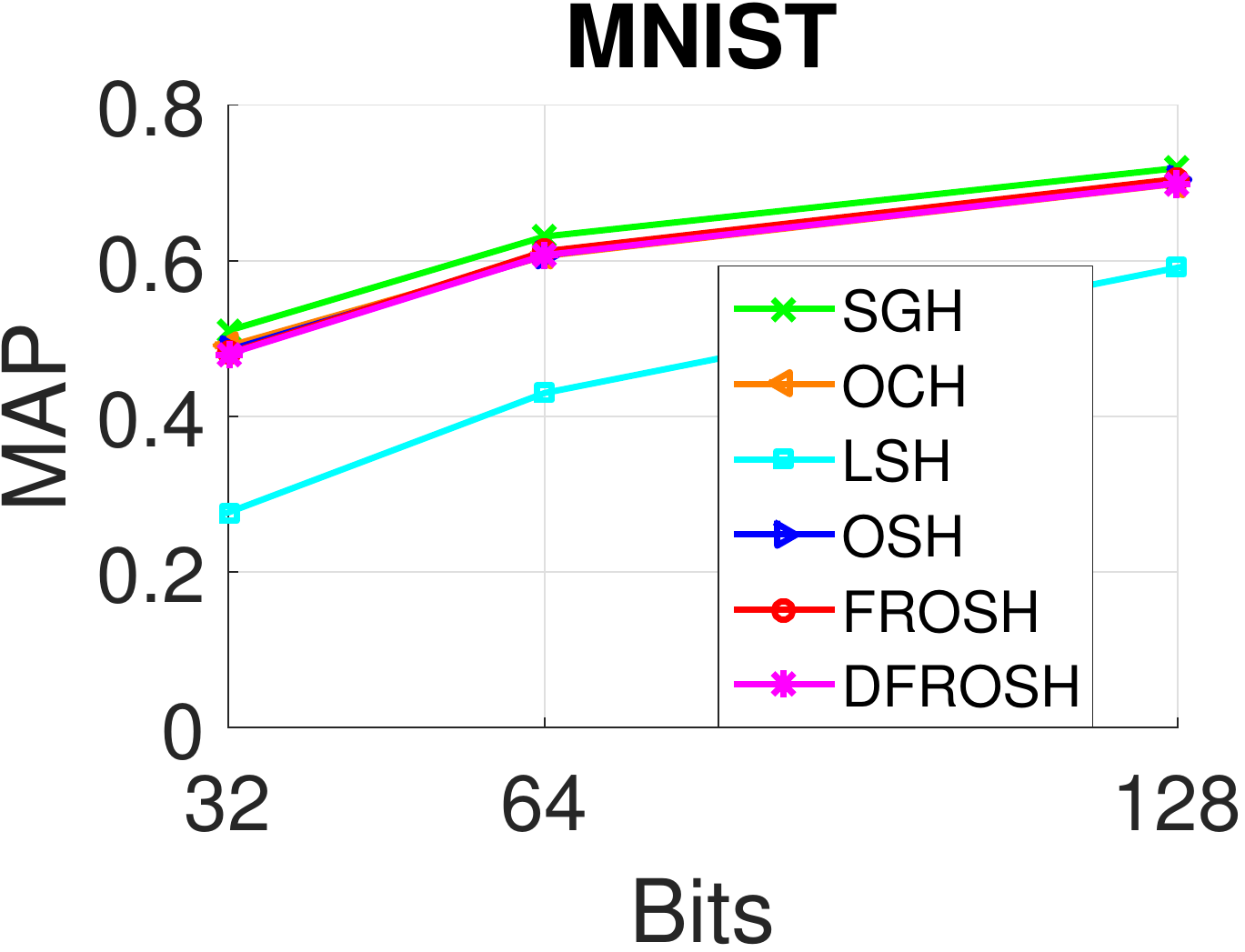}}
\subfigure
{\includegraphics[width=.24\textwidth]{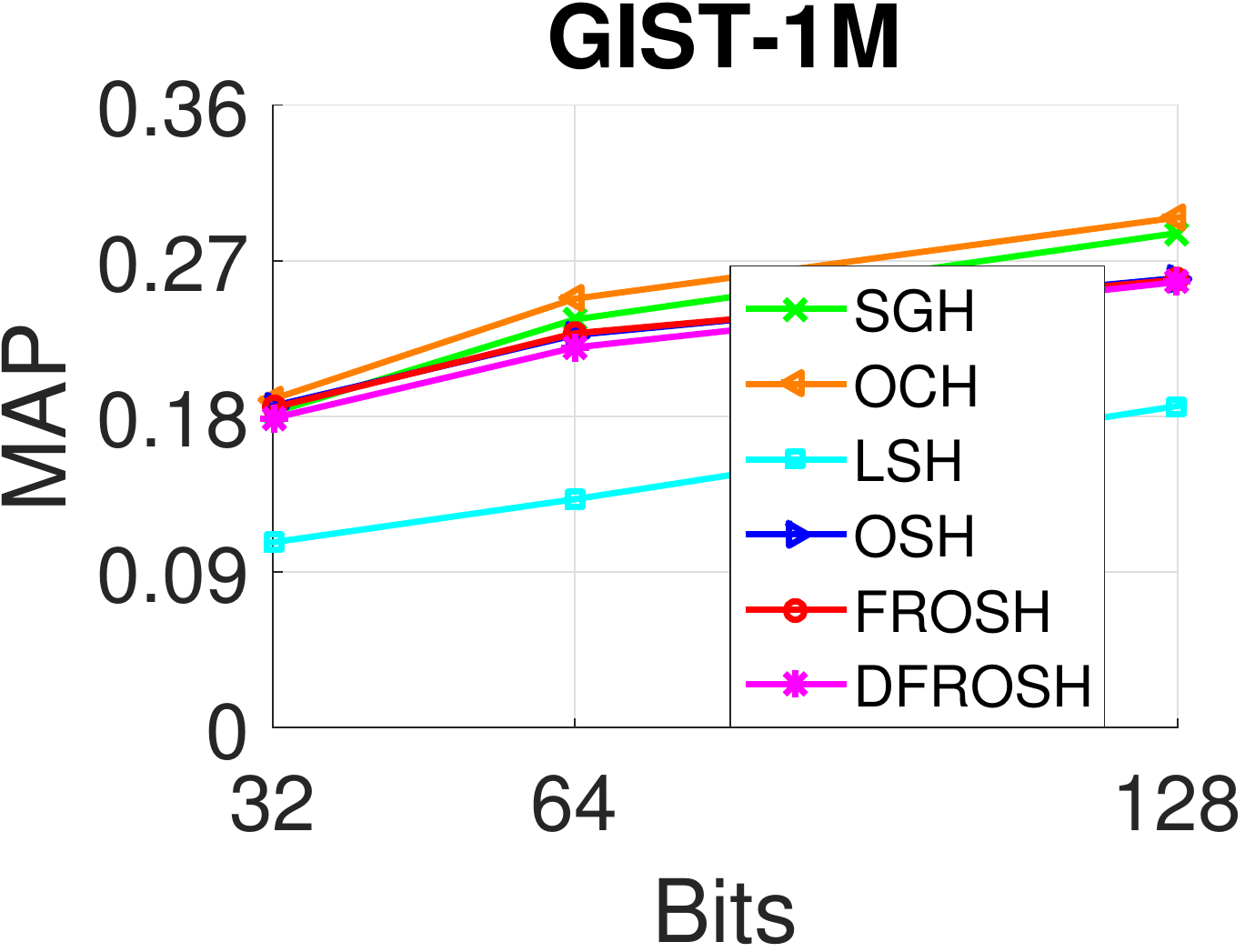}}
\subfigure
{\includegraphics[width=.24\textwidth]{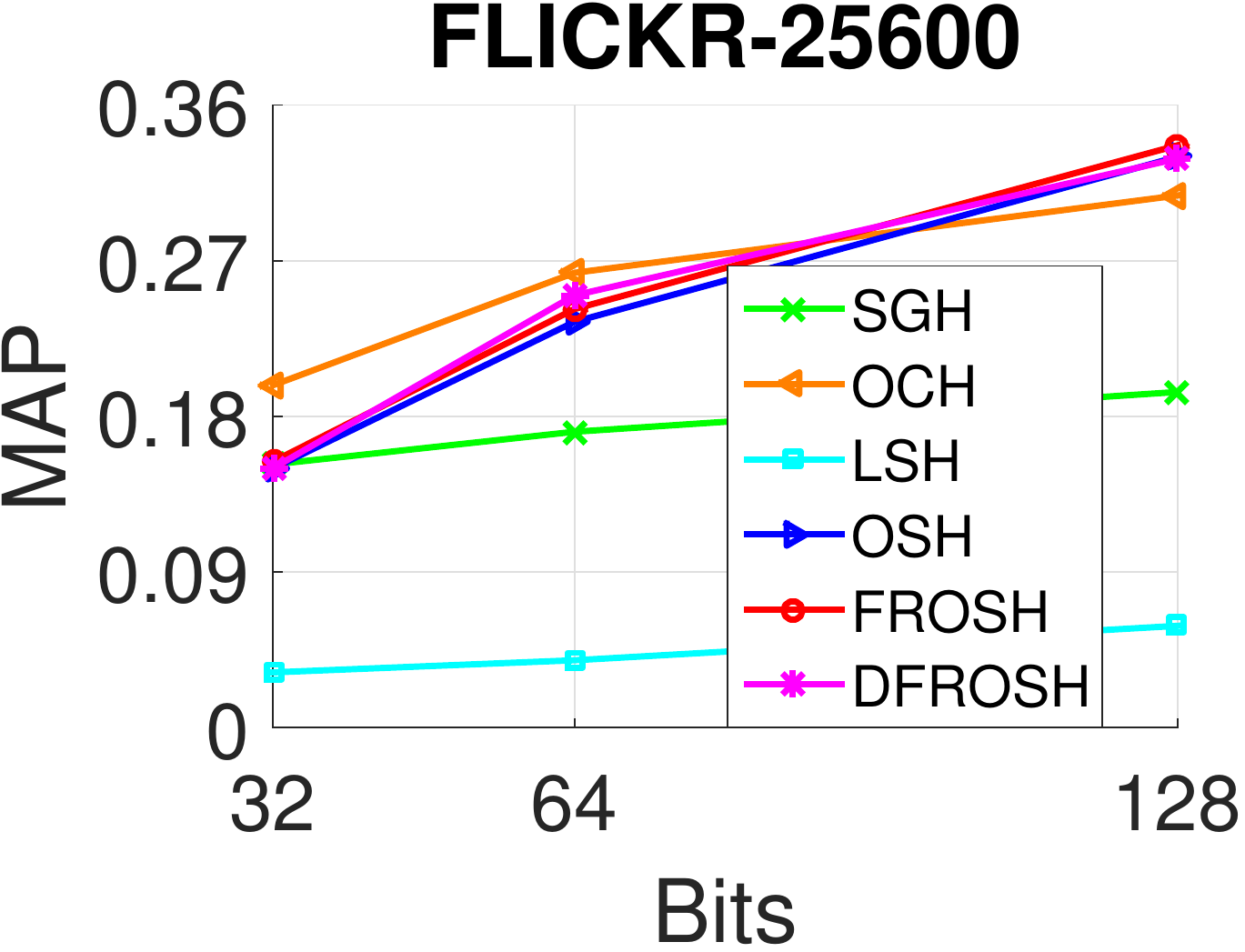}}
\caption{MAP comparisons at different code lengths.}
\label{fig:batch}
\end{figure*}

\subsection{Numerical Comparisons of FD and FFD}
Following the same setting of~\cite{liberty2013simple}, we construct $\A=\P\boldsymbol\Lambda\U+\Z/\gamma$, where $\P\in\R^{n\times k}$ is the signal coefficient matrix such that $P_{ij}\sim\mathcal{N}(0,1)$, $\boldsymbol\Lambda\in\R^{k\times k}$ is a square diagonal matrix with the diagonal entry being $\Lambda_{ii}=1-(i-1)/k$, which linearly diminishes the singular values, $\U\in\R^{k\times d}$ defines the signal row space with $\U\U^T=\I_d$,  $\Z\in\R^{n\times d}$ is the Gaussian noise, i.e., $Z_{ij}\sim\mathcal{N}(0,1)$, $\gamma$ is the parameter to control the effect of the noise and the signal, $k$ is the length of the controlling signal.  Usually, $k\ll d$.  We set both $k$ and $\gamma$ to $10$ as those in~\cite{liberty2013simple}.

In the evaluation, we vary the sketching size $\l$ from $\{16, 32, 64, 100, 128, 200, 256\}$, the dimension of the data $d$ from $\{64, 128, 256, 512\}$, and the number of the data $n$ from $\{${50,000}, {100,000}, {200,000}, {500,000}, {1,000,000}$\}$.  For FFD, we test the effect of $m$ by setting $m=\tau d$, where $\tau=\{1, 2, 4, 8\}$.  The relative error and the time cost are measured to evaluate the performance of FD and FFD under different settings, where the relative error is defined by $\|\A^T\A-\B^T\B\|_2/\|\A\|_F^2$ and $\B$ is the sketched matrix. 

Figure~\ref{fig:sketchingaccuracy} shows the relative errors and Figure~\ref{fig:sketchingtime} records the time cost for the compared algorithms, respectively.  The results show that 
\begin{compactitem}
\item FFD attains comparable accuracy to FD but enjoys much lower time cost compared with FD.  From Fig.~\ref{fig:syn_acc_vl}, the sketching precision increases gradually as the sketching size $\l$ increases while from Fig.~\ref{fig:syn_time_vl}, the time cost of FFD is significantly less than that of FD.  Moreover, the performance of FFD is insensitive to $m$.  Though the relative errors increase slightly when $m$ increases, the time cost can be further reduced in a certain magnitude.

\item The sketching errors decrease gradually with the increase of the number of data $n$ while the time cost scales linearly with $n$; see Fig.~\ref{fig:syn_acc_vn} and Fig.~\ref{fig:syn_time_vn} for the detailed results.

\item The sketching errors of both FD and FFD increase with the increase of the dimension $d$ while FFD is slightly worse than FD.  This follows our intuition and is in line with the observation in~\cite{liberty2013simple} because increasing $d$ may increase the intrinsic properties of data.  The time cost of both FD and FFD also scales linearly with $d$, but FFD costs significantly less time than FD.  The observations can be examined in Fig.~\ref{fig:syn_acc_vd} and Fig.~\ref{fig:syn_time_vd}.

\item In sum, the above observations of FFD are consistent with the theoretical results in Lemma~\ref{thm:FFD} while the observations of FD are in line with the results in~\cite{liberty2013simple}.  They demonstrate the the advantages of FFD over FD for online sketching hashing. 
\end{compactitem}

\subsection{Comparison of Online and Batch-trained Algorithms} \label{sec:comosh}
We compare our proposed FROSH and DFROSH with the following baseline algorithms: 
\begin{compactitem}
\item OSH~\cite{leng2015online}: our basic online sketching algorithm; 
\item {LSH~\cite{charikar2002similarity}: online sketching algorithm without training, i.e., conducting the hash projection function via a random Gaussian matrix;} 
\item Scalable Graph Hashing (SGH)~\cite{jiang2015scalable}: {a leading batch-trained algorithm which conducts graph hashing to efficiently approximate the graphs; } and 
\item Ordinal Constraint Hashing (OCH)~\cite{liu2017ordinal}: {another leading batch-trained algorithm that learns the optimal hashing functions from a graph-based approximation
to embed the ordinal relations}. 
\end{compactitem} 
and conduct on four benchmark real-world datasets: 
\begin{compactitem}
\item CIFAR-10~\cite{krizhevsky2009learning}: a collection of {60,000} images in {10} classes with each class of {6,000} images.  The GIST descriptors are employed to represent each image, which yields a $512$-dimensional data.  
\item MNIST~\cite{chang2011libsvm}: a collection of {70,000} images in {10} classes with each class of {7,000} images represented by $784$-dimensional features. 
\item GIST-1M~\cite{jegou2011product}: a collection of one million $960$-dimensional GIST descriptors.  
\item FLICKR-25600~\cite{yu2014circulant}: a collection of {100,000} images sub-sampled from web images with each image represented by a {25,600}-dimensional vector under instance normalization.  
\end{compactitem}
We follow the same measurement of~\cite{leng2015online}. Specifically, given a query, the returned point is set to its true neighbor if this point lies in the top 2\% closest points to the query, where the Euclidean distance is applied in the original space without hashing.  To test the accuracy performance of hashing, for each query, all data points are processed by hashing and ranked according to their Hamming distances to the query.  In the online algorithms, OSH, FROSH and DFROSH, we set the sketching size $\l$ to $2r$, where $r$ is the code length assigned from $\{32, 64, 128\}$.  $m$ is empirically set to $4d$ for FROSH and DFROSH, respectively.  In DFROSH, the data are evenly distributed in $5$ machines.  The setting of batch-trained algorithms SGH and OCH follows that in~\cite{jiang2015scalable, liu2017ordinal}. We measure precision-recall curves and the mean average precision (MAP) score of the compared algorithms, where the precision is computed via the ratio of true neighbors among all the returned points and the recall is computed via the ratio of the true neighbors among all ground truths while the precision-recall curve is obtained through changing the number of the returned points, and MAP records the area under the curve of precision-recall, which can be computed by averaging many precision scores evenly spaced along the recall to approximate such area. We have also divided the data into $10$ parts, and run the algorithms and record the performance part after part sequentially. Each part corresponds to a round.

Figure~\ref{fig:map} shows the MAP scores at different rounds with 32, 64 and 128 bits codes for online algorithms LSH, OSH, and FROSH. The comparison indicates that
\begin{compactitem}
\item On all datasets, it is apparent that our proposed FROSH performs
as accurately as OSH and outperforms LSH with a large margin. 
\item OSH and FROSH can stably improve the MAP scores when receiving more data, which demonstrates that a successful adaption to the data variations has been achieved.
\end{compactitem}

Figure~\ref{fig:batch} also includes the online algorithm DFROSH, and reports MAP scores of all compared algorithms with respect to different code lengths.  The results show that 
\begin{compactitem}
\item LSH yields the poorest performance because it does not learn any information from the data.  Meanwhile, OCH attains the best performance in most cases.  This indicates that the time cost of OCH is deserved.  
\item Our proposed FROSH and DFROSH attain comparable performance with OSH and even beat the two leading batch-trained algorithms in FLICKR-25600, i.e., the data with extremely high dimension. 
\end{compactitem}

\begin{figure*}[htbp]
\centering
\subfigure
{\includegraphics[width=.24\textwidth]{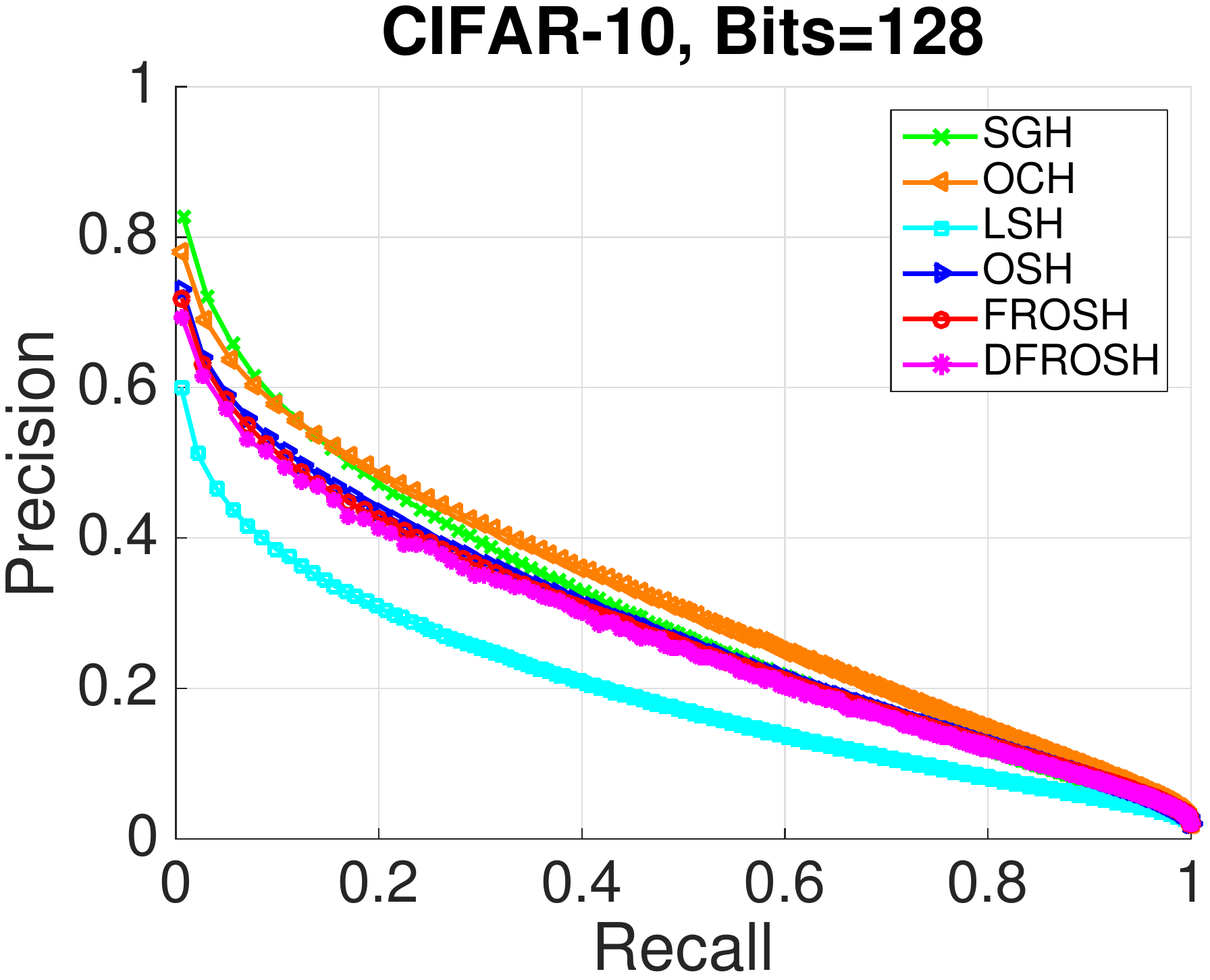}}
\subfigure
{\includegraphics[width=.24\textwidth]{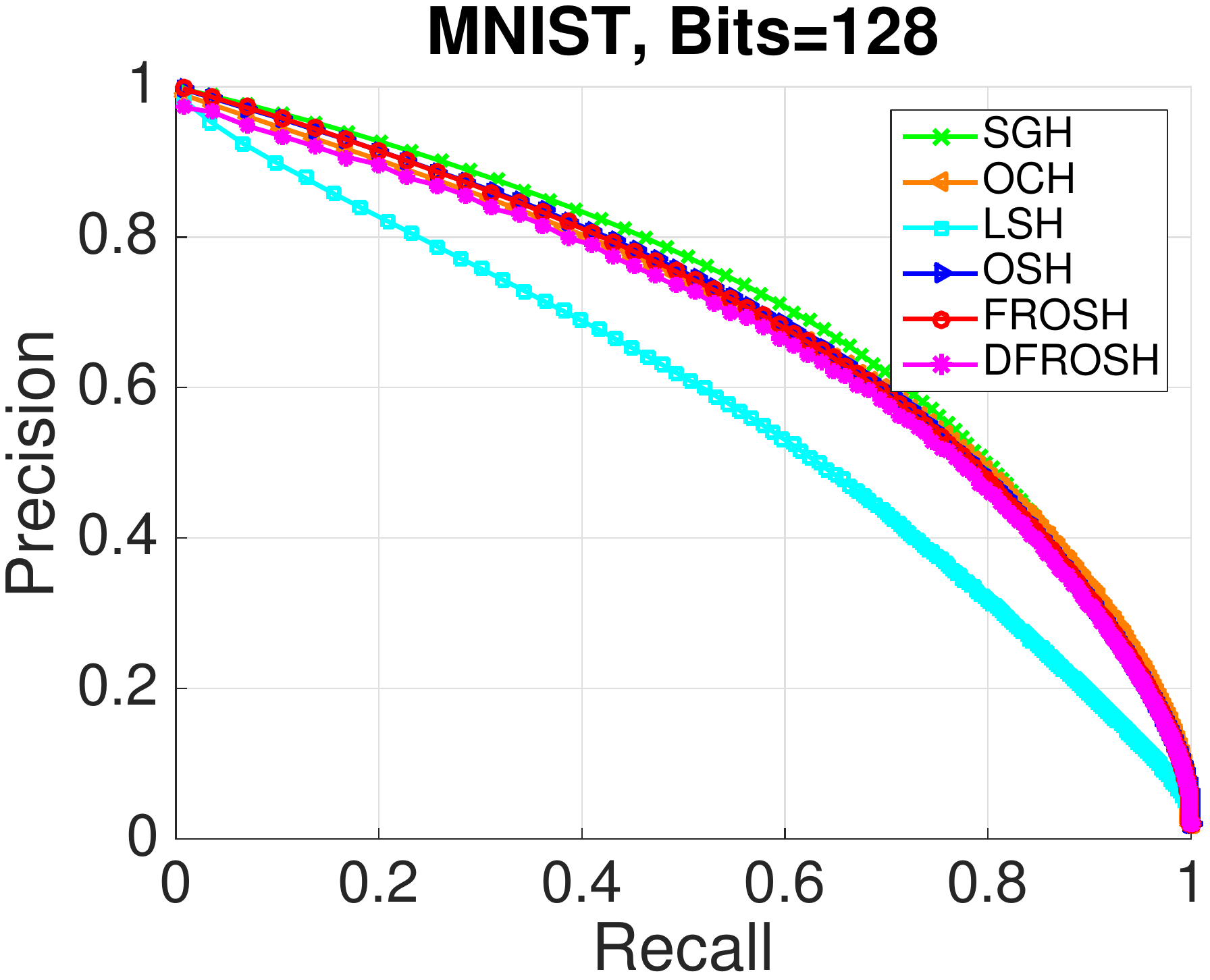}}
\subfigure
{\includegraphics[width=.24\textwidth]{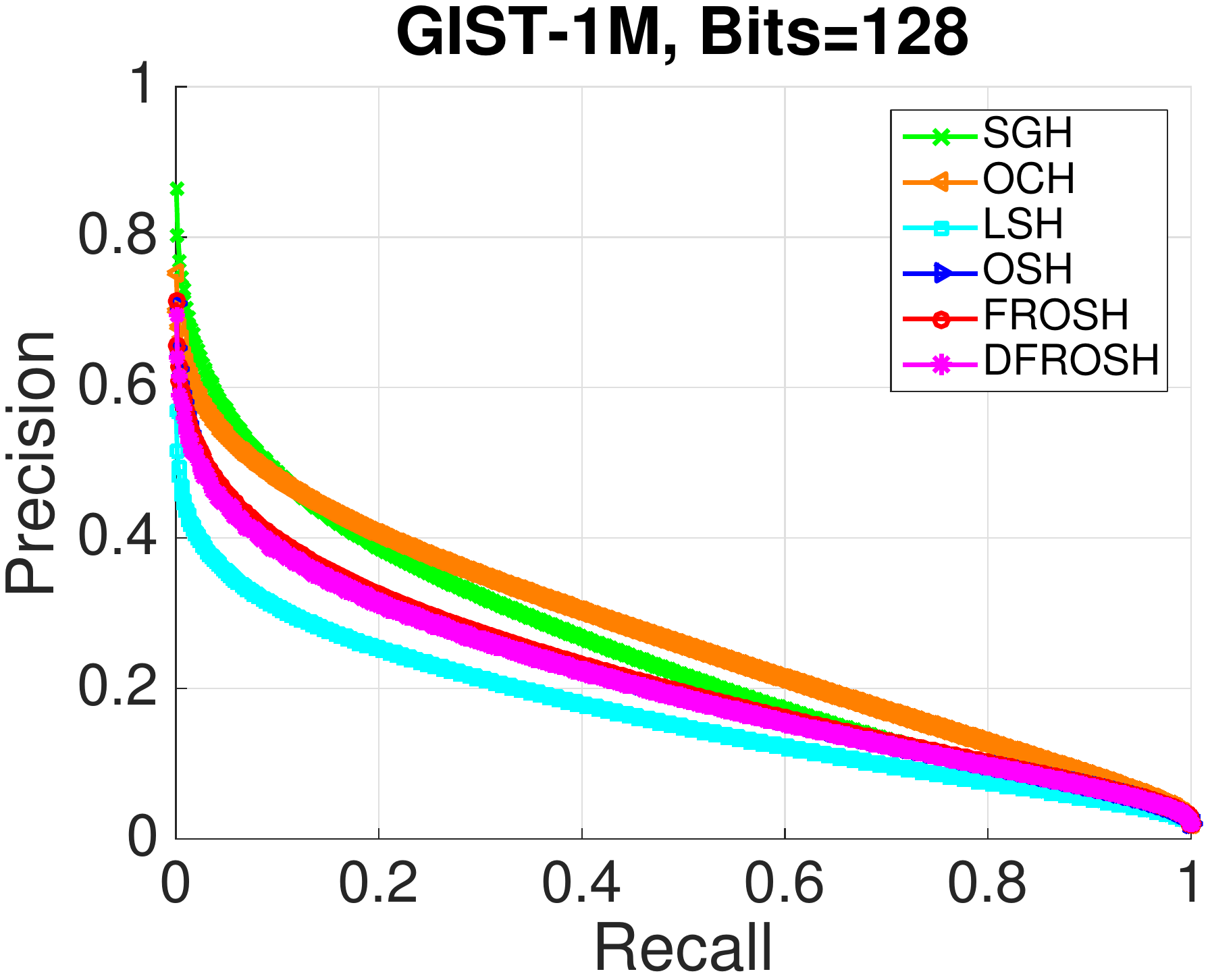}}
\subfigure
{\includegraphics[width=.24\textwidth]{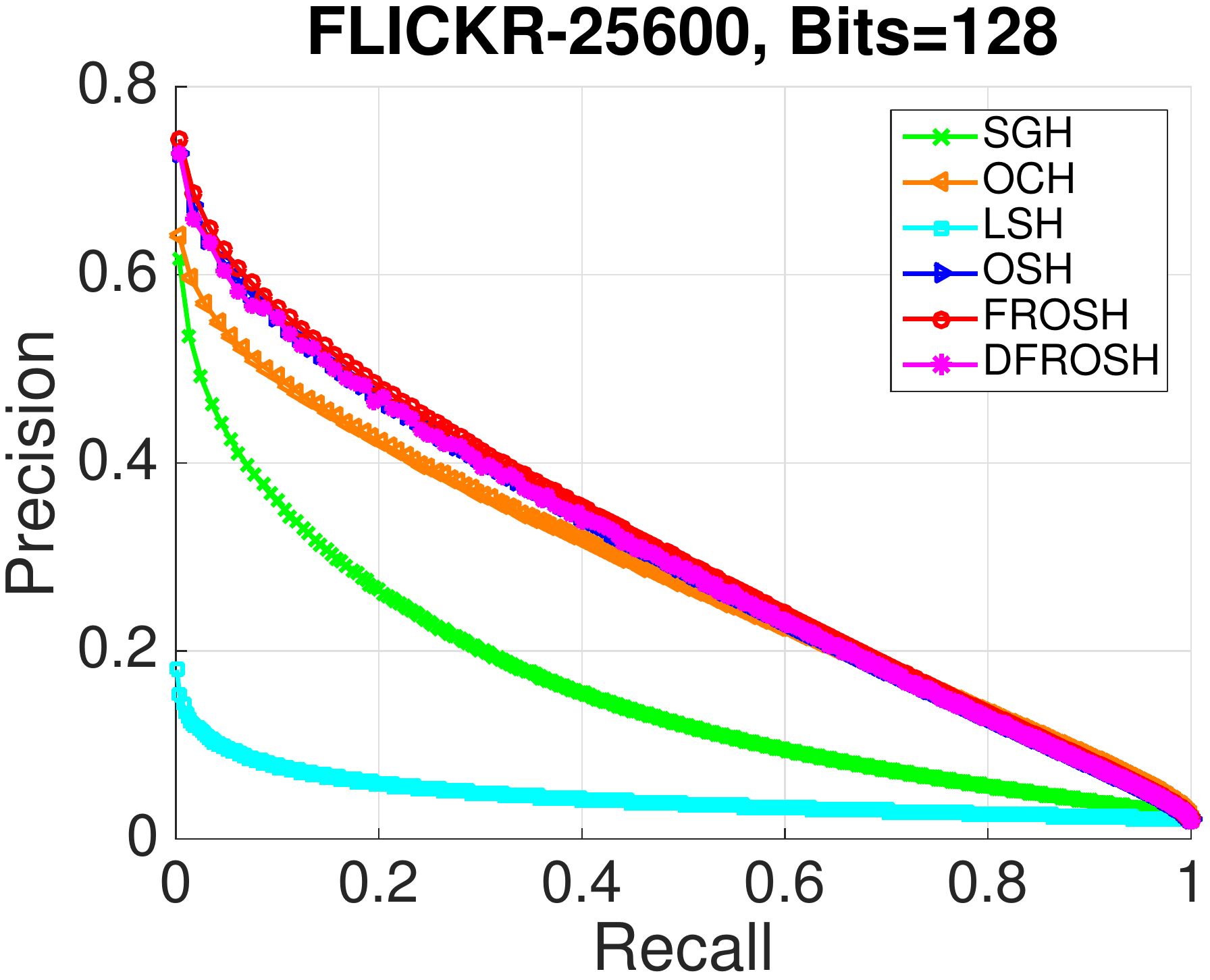}}
\caption{Precision-Recall comparisons for the code length of 128.}
\label{fig:recall_precision}
\end{figure*}

Figure~\ref{fig:recall_precision} presents the precision-recall curves of the compared algorithms with the code length of 128. The curves of OSH, FROSH, and DFROSH almost overlap exhibit competitive or better performance compared with other methods. 

Table~\ref{tab:hashingtime} records the time cost of all compared algorithms.  The results show that 
\begin{compactitem}
\item OCH costs the most time while SGH and OSH yielding similar time cost on CIFAR-10, MNIST, GIST-1M.  For FLICKR-25600, when the number of dimension is large, SGH costs significantly more time than OSH.  These observations are in line with previous work in~\cite{charikar2002similarity,jiang2015scalable,leng2015online,liu2017ordinal}. 
\item FROSH yields significantly less training time than OSH, only one-twentieth to one-tenth of OSH.
\item DFROSH is the most efficient algorithm.  Since it runs on five machines simultaneously, its time cost is around one-fifth of FROSH.  Overall, it speeds up $40\sim 300$ times compared with the batch-trained algorithms. 
\end{compactitem}

In a word, our proposed FROSH and DFROSH can adapt hash functions to new coming data and enjoy superior training efficiency, i.e., single pass, low computational cost and low memory cost.  
 



\begin{table}[htbp]
\caption{The training cost (in sec.) of SGH, OCH, OSH, FROSH, and DFROSH on four real-world datasets w.r.t. code lengths.}
\label{tab:hashingtime}
\begin{center}
\begin{tabular}{c||c||c|c|c}
\toprule
\textbf{Dataset} & \textbf{Algorithm} & $32$ bits & $64$ bits & $128$ bits \\
\midrule
\multirow{5}{*}{CIFAR-10} 
               & SGH & 7.83 & 11.35 & 19.49\\  \cline{2-5} 
               & OCH & 26.89 & 26.95 & 27.49\\  \cline{2-5} 
               & OSH & 7.78 & 11.88 & 22.09\\  \cline{2-5} 
               & \textbf{FROSH} & \textbf{0.63} & \textbf{0.94}  & \textbf{2.11}\\
               \cline{2-5} 
               & \textbf{DFROSH} & \textbf{0.13} & \textbf{0.20}  & \textbf{0.44}\\
\hline
\hline
              \multirow{5}{*}{MNIST} 
               & SGH & 10.47 & 14.59 & 23.47\\  \cline{2-5} 
               & OCH & 40.45 & 40.49 & 41.10\\  \cline{2-5} 
               & OSH & 13.25 & 18.93 & 30.75\\  \cline{2-5} 
               & \textbf{FROSH} & \textbf{1.17} & \textbf{1.49} & \textbf{2.56} \\
               \cline{2-5} 
               & \textbf{DFROSH} & \textbf{0.24} & \textbf{0.31}  & \textbf{0.53}\\
\hline
\hline
              \multirow{5}{*}{GIST-1M} 
               & SGH & 231 & 275& 290\\  \cline{2-5} 
               & OCH & 1,042 & 1,089 & 1,192\\  \cline{2-5} 
               & OSH & 228 & 331 & 520\\  \cline{2-5} 
               & \textbf{FROSH} & \textbf{21} & \textbf{27} & \textbf{45} \\
               \cline{2-5} 
               & \textbf{DFROSH} & \textbf{4.3} & \textbf{5.6}  & \textbf{9.7}\\
\hline
\hline



               & SGH & 3,032 & 3,541 & 4,903\\  \cline{2-5} 
           FLICKR-    & OCH & 4,981 & 5,300 & 5,441\\  \cline{2-5} 
           25600    & OSH & 679 & 1,283& 2,570\\  \cline{2-5} 
               & \textbf{FROSH} & \textbf{72} & \textbf{92} & \textbf{134} \\
               \cline{2-5} 
               & \textbf{DFROSH} & \textbf{15.6 } & \textbf{19.4}  & \textbf{29.2}\\
\bottomrule
\end{tabular}
\end{center}
\end{table} 

\section{Conclusions}
In this paper, we overcome the inefficiency of the OSH algorithm and propose the FROSH algorithm and its distributed implementation.  We provide rigorous theoretical analysis to guarantee the sketching precision of FROSH and DFROSH and their training efficiency.  We conduct extensive empirical evaluations on both synthetic and real-world datasets to justify our theoretical results and practical usages of our proposed FROSH and DFROSH. 




\bibliographystyle{abbrv}
\bibliography{egbibc}

\end{document}